\documentclass[lettersize,journal]{IEEEtran}
\usepackage{amsmath,amsfonts}
\usepackage{textcomp}
\usepackage{stfloats}
\usepackage{url}
\usepackage{verbatim}
\usepackage{graphicx}
\usepackage[nocompress]{cite}
\usepackage{subfigure}
\usepackage{amsmath,amssymb,amsfonts}
\usepackage{algorithmic}
\usepackage{multirow}
\usepackage{amssymb}
\usepackage{pifont}
\usepackage{makecell}
\usepackage{xcolor}
\usepackage{pifont}

\usepackage{indentfirst}
\usepackage{epstopdf}
\usepackage{tabularx}
\usepackage{times}
\usepackage{epsfig}
\usepackage{color}
\usepackage{overpic}
\usepackage{wrapfig,lipsum,booktabs}
\usepackage{ragged2e}
\definecolor{ballblue}{rgb}{0.13, 0.67, 0.8}
\usepackage[colorlinks,bookmarksnumbered,bookmarksopen]{hyperref}
\usepackage{bm}

\usepackage[ruled,linesnumbered,lined,boxed,commentsnumbered]{algorithm2e}
\definecolor{ballblue}{rgb}{0.13, 0.67, 0.8}

\ifdefined \GramaCheck
  \newcommand{\CheckRmv}[1]{}
  \newcommand{\figref}[1]{Figure 1}%
  \newcommand{\tabref}[1]{Table 1}%
  \newcommand{\secref}[1]{Section 1}
  \renewcommand{\eqref}[1]{Equation 1}
\else
  \newcommand{\CheckRmv}[1]{#1}
  \newcommand{\figref}[1]{Fig.~\ref{#1}}%
  \newcommand{\tabref}[1]{Table~\ref{#1}}%
  \newcommand{\secref}[1]{Sec.~\ref{#1}}
  \renewcommand{\eqref}[1]{Equation~(\ref{#1})}
\fi

\begin{document}
\title{BCEdge: SLO-Aware DNN Inference Services \\with Adaptive Batching on Edge Platforms}

\author{Ziyang Zhang, \IEEEmembership{Student Member, IEEE}, Huan Li, \IEEEmembership{Senior Member, IEEE}, Yang Zhao, \\
\IEEEmembership{Senior Member, IEEE}, Changyao Lin, and Jie Liu, \IEEEmembership{Fellow, IEEE}

\thanks{This work is partly supported by the National Key R\&D Program of China under Grant No. 2021ZD0110905, and An Open Competition Project of Heilongjiang Province, China, on Research and Application of Key Technologies for Intelligent Farming Decision Platform, under Grant No. 2021ZXJ05A03.}
\thanks{Ziyang Zhang is with the Harbin Institute of Technology, Harbin, Heilongjiang 150006 China (e-mail: zhangzy@stu.hit.edu.cn). \\
Huan Li is with the Harbin Institute of Technology, Shenzhen, Guangdong 518071 China (e-mail: huanli@hit.edu.cn). \\
Yang Zhao is with the Harbin Institute of Technology, Shenzhen, Guangdong 518071 China (e-mail: yang.zhao@hit.edu.cn). \\
Changyao Lin is with the Harbin Institute of Technology, Harbin, Heilongjiang 150006 China (e-mail: lincy@stu.hit.edu.cn). \\
Jie Liu is with the Harbin Institute of Technology, Shenzhen, Guangdong 518071 China (e-mail: jieliu@hit.edu.cn).}}
\maketitle

\begin{abstract}
As deep neural networks (DNNs) are being applied to a wide range of edge intelligent applications, it is critical for edge inference platforms to have both high-throughput and low-latency at the same time.
Such edge platforms with multiple DNN models pose new challenges for scheduler designs.
First, each request may have different service level objectives (SLOs) to improve quality of service (QoS). Second, the edge platforms should be able to efficiently schedule multiple heterogeneous DNN models so that system utilization can be improved.
To meet these two goals, this paper proposes BCEdge, a novel learning-based scheduling framework that takes adaptive batching and concurrent execution of DNN inference services on edge platforms.
We define a utility function to evaluate the trade-off between throughput and latency.
The scheduler in BCEdge leverages maximum entropy-based deep reinforcement learning (DRL) to maximize utility by 1) co-optimizing batch size and 2) the number of concurrent models automatically. 
Our prototype implemented on different edge platforms shows that the proposed BCEdge enhances utility by up to 37.6\% on average, compared to state-of-the-art solutions, while satisfying SLOs.
\end{abstract}


\section{Introduction}\label{sec:introduction}
\IEEEPARstart {M}odel inference service systems deployed on cloud servers typically provide multiple trained deep neural networks (DNNs) for users.
These systems are usually multi-tenant, meaning hosting one or more model instances per DNN model to serve multiple inference applications, while making better use of the abundant computing resources of servers.
For instance, the multi-instance GPU (MIG) in NVIDIA Ampere architecture enables the partitioning of a single NVIDIA A100 GPU into up to seven independent GPU instances that can run concurrently.
In this way, the GPU achieves up to 7$\times$ utilization with guaranteed quality of service (QoS).
Furthermore, a single inference request often leads to inefficient utilization.
Therefore, prior works~\cite{crankshaw2017clipper,olston2017tensorflow,Triton} batch requests to better exploit the parallelism of GPUs.
Batching refers to aggregating arriving requests into a batch within a given time window, and DNN service systems process the entire batch at a particular time, thereby improving throughput (e.g., requests per second, rps).
In batch systems, throughput can be improved by increasing batch size. The more requests in a batch, the longer the waiting time to be processed, latency, therefore, is inevitably increased.

Increasing computational and memory capabilities open a new opportunity to deploy model inference systems on edge accelerators (e.g., graphics processing unit (GPU), tensor processing unit (TPU), and vision processing unit (VPU), etc.).
This emerging computing paradigm provides guarantee for edge intelligent applications~\cite{deng2020edge} with low-latency requirements, including the object detection in autonomous driving~\cite{feng2020deep}, recommendation systems in smartphones~\cite{xie2022contrastive}, and the metaverse in wearables~\cite{liu2020collabar}, etc. 
On the other hand, various lightweight techniques~\cite{han2015deep} (such as pruning, compression, quantization, knowledge distillation, etc.) for DNN models enable batching and concurrent inference of multiple model instances at the edge.

\begin{figure} 
\vspace{0pt}
\setlength{\abovecaptionskip}{0pt}
\setlength{\belowcaptionskip}{0pt}
\centering
\subfigure[Throughput (rps)]{
\begin{minipage}[b]{0.465\linewidth}
\includegraphics[width=1\linewidth]{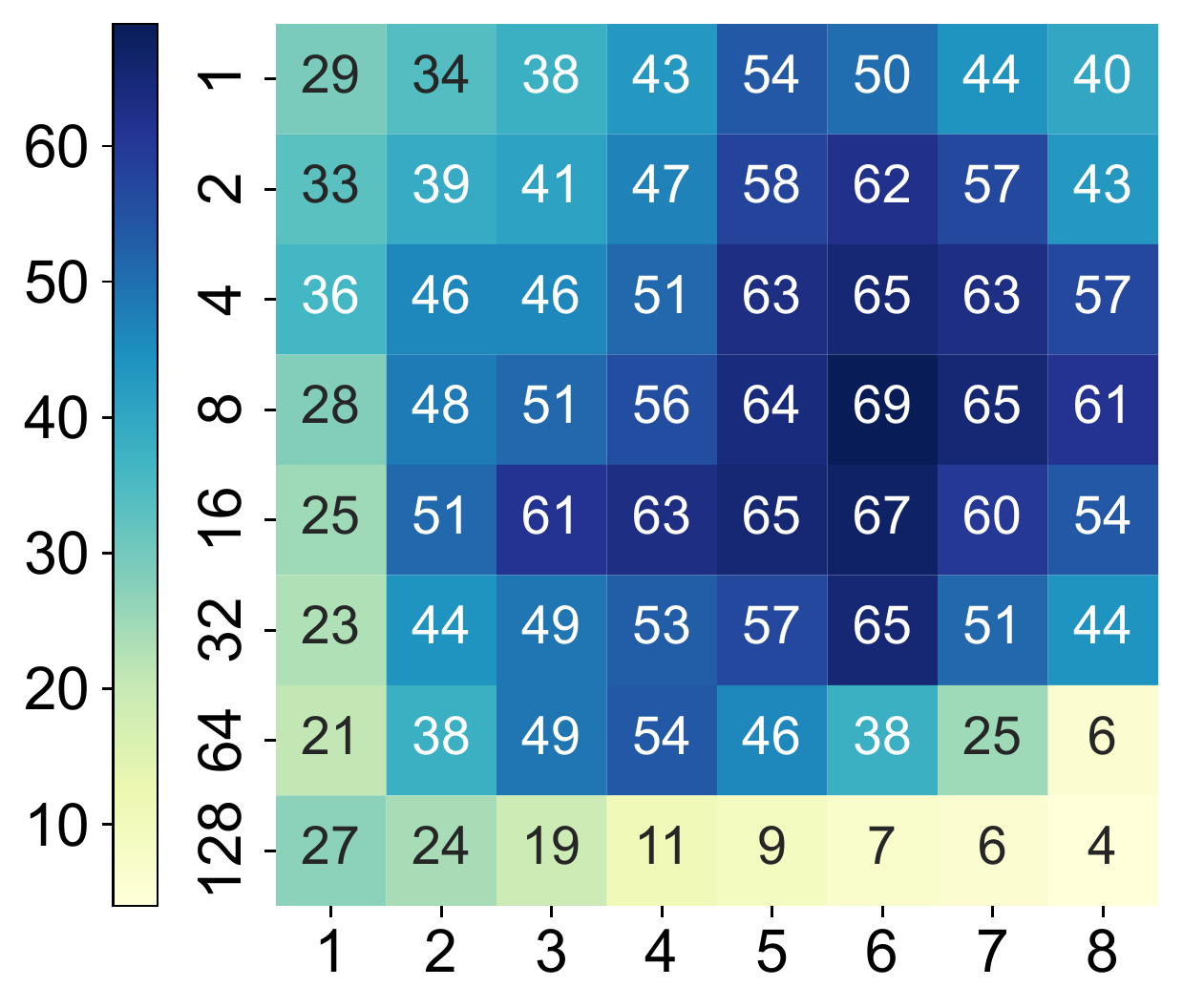}
\end{minipage}}
\subfigure[Latency (ms)]{
\begin{minipage}[b]{0.48\linewidth}
\includegraphics[width=1\linewidth]{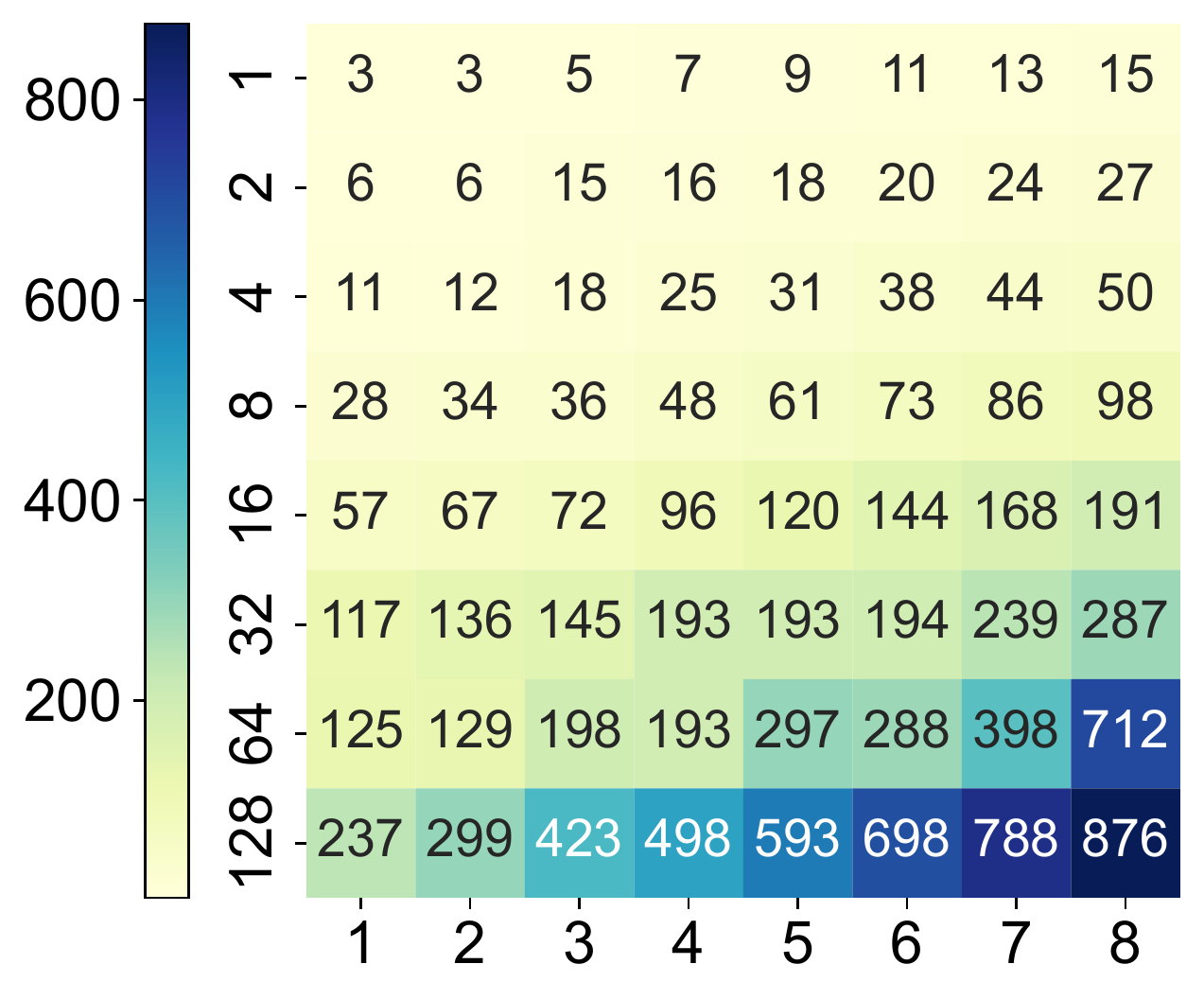}
\end{minipage}}
\caption{The effects of batching and concurrent inference on (a) system throughput and (b) end-to-end latency. Throughput is measured as requests-per-second (rps). The x-axis represents the number of concurrent models, and the y-axis represents batch size. We use YOLO-v5~\cite{YOLOv5} on NVIDIA Xavier NX edge platform with 8GB RAM.}
\label{observation}
\vspace{-0.2cm} 
\end{figure}

To better understand the performance implications of batching and concurrent inference, we perform an experimental study, using YOLO-v5~\cite{YOLOv5} on NVIDIA Xavier NX edge platform.
Due to resource constraints on edge platforms, we leverage TensorRT~\cite{TensorRT} to accelerate the original DNN models. 
Fig.~\ref{observation} reports the throughput and latency with various batch sizes and number of concurrent models. 
We have the following critical observations that motivate this work:
\emph{both batch size and number of concurrent models affect throughput and latency, but larger batch size or number of models is not always better.} 
Fig.~\ref{observation} illustrates that higher-throughput and lower-latency appear in moderate batch size and number of concurrent models.
Due to resource contention caused by model interference, excessive batch size and number of concurrent models significantly reduce throughput and increase latency or even cause memory overflow, especially when batch size and number of concurrent models are large, e.g., batch size is 128, model number is 8.
Therefore, it is crucial for the scheduler in DNN service system to 1) trade off throughput and latency for optimal performance
and 2) accurately predict the interference between models.

Designing such a GPU-based inference servers must address different challenges from batching- and concurrent-oriented processing.
First, inference requests have service level objective (SLO) to achieve quality of service (QoS) with low-latency.
Second, edge platforms usually deploy multiple heterogeneous DNN models to improve throughput and resource utilization. 
Therefore, it is critical to design an efficient scheduler to achieve the optimization goal for both batching and concurrent inference on edge platforms. 
Conventional heuristic-based methods are inefficient for multi-objective optimization~\cite{tuli2020dynamic}. 
In contrast, deep reinforcement learning (DRL) combines the powerful representation of deep learning with the adaptive property of reinforcement learning, which is capable of efficiently solving the above problems. 
Therefore, we leverage DRL to transform a multi-objective (i.e., throughput and latency) problem into a scheduling problem of batch size and number of concurrent models.

Motivated by the above observation, we propose BCEdge, a learnable, adaptive, and multi-tenant scheduling framework for SLO-aware DNN inference services.
BCEdge aims to automatically find a global optimum by adjusting both batch size and number of concurrent models for throughput and latency tradeoffs.
The search space for scheduling becomes two-dimensional with batch size and number of concurrent models, unlike the prior work with one-dimensional (i.g., batch size) searches.
The batching-concurrent scheduling can significantly improves the SLO-preserved throughput.
For each DNN model, its computational properties are measured and registered into BCEdge. 
Based on the profile information of each DNN model, the maximum entropy reinforcement learning-based scheduler in BCEdge automatically adjust batch size and number of concurrent models, while maintaining the SLO.
Furthermore, BCEdge leverages a lightweight neural network (NN)-based interference prediction model to reduce the impact of concurrent inference.

Table~\ref{related} provides a summarized comparison of our work to related DNN service frameworks.
All previous studies are capable of adaptively adjusting batch size at runtime, either automatically or manually, to achieve higher-throughput.
As more DNN inference services are consolidated into edge-based GPU servers, although some previous studies provided multiple heterogeneous DNN models for multi-tenancy, these works did not enable scheduling multiple instances of the same model in DNN service system, which becomes more important to make full use of edge servers with constantly increasing computing resources.
Regarding inter-model resource contention among multi-tenants and requests with SLO, accurately predicting interference and guaranteeing SLO can better guide the scheduler to automatically adjust batch size and number of concurrent models to achieve moderate system throughput and reduce latency. Only TF-Serving~\cite{olston2017tensorflow} considers model interference, and the scheduler uses hedged backup requests to mitigate latency spikes caused by inter-request or -model interference.
On the other hand, except for Clipper~\cite{crankshaw2017clipper} and DeepRT~\cite{yang2021deeprt}, none of other previous works consider requests with strict time constraints (i.e., SLO).
Importantly, our study addresses all challenges, executing multi-instance concurrently, guaranteeing SLO, and predicting potential interference among multi-model.
\begin{table}
\scriptsize
\setlength{\abovecaptionskip}{0pt}
\setlength{\belowcaptionskip}{0pt}
    \caption{Comparison with prior work}
    \label{related}
    \centering
    \begin{tabular}{cccccc} \hline
    	\textbf{\makecell[c]{Service \\ Framework}}  & \textbf{\makecell[c]{Adaptive \\ Batching}} & \textbf{\makecell[c]{Concurrent \\ Instance}} & \textbf{\makecell[c]{Multi \\ Model}} & \textbf{\makecell[c]{Interference \\ Prediction}} & \textbf{\makecell[c]{SLO \\ Aware}} \\ \hline
     
        TF-Serving~\cite{olston2017tensorflow} & \textcolor{teal}{\checkmark} & \textcolor{purple}{\ding{55}} & \textcolor{teal}{\checkmark}  & \textcolor{teal}{\checkmark} & \textcolor{purple}{\ding{55}} \\
     
    	Triton~\cite{Triton}  & \textcolor{teal}{\checkmark} & \textcolor{teal}{\checkmark} & \textcolor{teal}{\checkmark}  & \textcolor{purple}{\ding{55}}  & \textcolor{purple}{\ding{55}} \\ 

    	DeepRT~\cite{yang2021deeprt} & \textcolor{teal}{\checkmark} & \textcolor{purple}{\ding{55}}  & \textcolor{teal}{\checkmark}  & \textcolor{purple}{\ding{55}} & \textcolor{teal}{\checkmark} \\
     
        Clipper~\cite{crankshaw2017clipper} & \textcolor{teal}{\checkmark} & \textcolor{purple}{\ding{55}} & \textcolor{teal}{\checkmark}  & \textcolor{purple}{\ding{55}} & \textcolor{teal}{\checkmark} \\  
        
        Prema~\cite{choi2020prema} & \textcolor{teal}{\checkmark} & \textcolor{purple}{\ding{55}} & \textcolor{teal}{\checkmark}  & \textcolor{purple}{\ding{55}} & \textcolor{purple}{\ding{55}} \\ 

        DVABatch~\cite{cui2022dvabatch} & \textcolor{teal}{\checkmark} & \textcolor{teal}{\checkmark}  & \textcolor{teal}{\checkmark}  & \textcolor{purple}{\ding{55}} & \textcolor{purple}{\ding{55}} \\

        \textbf{BCEdge (Ours)} & \textcolor{teal}{\checkmark} & \textcolor{teal}{\checkmark}  & \textcolor{teal}{\checkmark}  & \textcolor{teal}{\checkmark} & \textcolor{teal}{\checkmark} \\ \hline
    \end{tabular}
\end{table}

We evaluated the proposed DNN inference framework on edge platforms with three heterogeneous edge GPUs, using six DNN models in Table~\ref{model} that cover both CV and NLP applications, such as object detection, image classification as well as speech recognition.
The evaluation shows that the proposed scheduling technique with batching and concurrent model instance can improve the trade-off with SLO constraints by 37.6\%, compared to the state-of-the-art solutions.
The main contributions of this paper are as follows:
\begin{itemize}
\item
Through a motivational case study based on real-world batching and concurrent inference of DNN models on edge platforms, we demonstrate that the trade-off between throughput and latency should leverage both batching and concurrent inference.
\item 
We present BCEdge, a learnable scheduling framework with adaptive batching and concurrent model instance for inference service on edge platforms. The scheduler in BCEdge leverages the maximum entropy reinforcement learning to automatically adjust batch size and number of concurrent models to trade off throughput and latency.
\item
For accurate performance prediction, the lightweight NN-based prediction model with negligible overhead in BCEdge reduces the effect of interference among models for DNN concurrent inference.
\end{itemize}

The rest of the paper is organized as follows: Section~\ref{Related} presents related work. Section \ref{Model} describes system model and problem formulation. Section~\ref{Design} illustrates our framework design in detail. Section~\ref{Evaluation} reports experimental results. Section~\ref{Conclusion} concludes our work.

\section{Related work}\label{Related}
\subsection{Model-level DNN Inference Service}
Prior works treated the DNN model as an indivisible whole, and proposed a series of edge inference serving frameworks to provide the quality of DNN inference services~\cite{olston2017tensorflow,zhang2019mark,yang2021deeprt,cui2022dvabatch,choi2022serving,seo2021slo,romero2021infaas,258862,crankshaw2017clipper,ali2020batch}. 
Clipper~\cite{crankshaw2017clipper}, TensorFlow-Serving~\cite{olston2017tensorflow}, MArK~\cite{zhang2019mark}, DeepRT~\cite{yang2021deeprt}, and BATCH~\cite{ali2020batch} adopt the traditional adaptive batching that use time window for efficient DNN inference.
None of these existing frameworks offer concurrent operation of model instances to further improve throughput. 
There are also some prior works research on SLO-aware DNN inference service. 
Gpulet\cite{choi2022serving} leverages spatio-temporal sharing of computing resources for multiple heterogeneous DNN models with SLO constraints. 
Clockwork~\cite{258862} exploits predictable execution times to achieve tight request-level SLO.
INFaaS~\cite{romero2021infaas} reduces costs, better throughput, and fewer SLO violations by choosing an adequate variation of a model.
PSLO~\cite{seo2021slo} is a preempting SLO-aware scheduler based on minimum average expected latency for edge platforms, which aims to trade-off response time, system throughput, and SLO.
Different from the above works, we foucs on reducing the SLO violation rate caused by the interference of multi-model.
In addition, some edge inference frameworks involve privacy protection~\cite{liu2021leia, hou2021model} and edge-cloud collaborative~\cite{kim2020autoscale, zhang2021deep}, respectively. These works are orthogonal to BCEdge that can alleviate privacy and resource constraints.

\subsection{Operator-level DNN Inference Service}
There are some prior research on optimizing the operator scheduling of DNN models to improve the quality of model service~\cite{han2022microsecond,cui2021enable,choi2020prema,liu2022veltair}.
REEF\cite{han2022microsecond} apopts a parallel mechanism based on dynamic kernel padding to improve the overall throughput.
VELTAIR~\cite{liu2022veltair} proposed an adaptive operator-level compilation and scheduling to guarantee resource usage efficiency and reduce interference-induced performance loss for multi-tenant DNN services.
PREMA~\cite{choi2020prema} is a predictive multi-task scheduling algorithm for preemptible neural processing unit to meet high-throughput.
Abacus~\cite{cui2021enable} leverages overlap-aware latency prediction and deterministic scheduling of overlapped DNN operators that improves throughput while maintaining the QoS for multi-tenant DNN services.
Since BCEdge exploits the computing power of accelerators on edge platforms using batching and concurrent inference, these works are also orthogonal to BCEdge and can be combined together to enable even higher-throughput and lower-latency.

\subsection{Multi-tenant Scheduling on Edge Platforms}
Multi-tenant scheduling is more challenging due to resource constrained on edge platforms, compared with cloud computing.
TVW-RL~\cite{mondal2021scheduling} exploit various temporal resource usage patterns of time-varying workloads based on a deep reinforcement learning (DRL) approach, to improve utilization in real production traces.
Likewise, KaiS\cite{han2021tailored}, A3C-R2N2~\cite{tuli2020dynamic}, MILP~\cite{wang2020multi}, A3C-DO~\cite{zou2020a3c}, and  MFRL~\cite{shi2020mean} proposed different multi-agent reinforcement learning-based scheduling strategies in edge-cloud cluster, to optimize throughput, latency, energy consumption, cost, etc.
MCDS~\cite{tuli2021mcds} uses a tree-based search strategy and a DNN-based prediction model to optimize QoS in edge-cloud testbeds.
Similar to MILP\cite{wang2020multi}, DeEdge~\cite{meng2019online} proposed D-Deads, a distributed greedy scheduling algorithm with task-deadline in edge computing, which maximize throughput while minimizing latency.
Note that the above works only schedule individual tasks one by one, ignoring the benefits of batching and concurrent inference.
Inspired by these works, BCEdge can also be extended to an edge-cloud collaborative inference framework to optimize specific objectives.

\section{System Model and Problem formulation}\label{Model}
In this section, we first formulate system model, including request-, scheduling-, computing-and networking model. Next, we present the optimization problem formulation to show the trade-off between throughput and latency.

\subsection{System Model}\label{Interference}
\subsubsection{Request Model} 
We assume that the IoT devices (e.g., cameras, drones, smartphones, etc.) share the computing resources of edge platforms.
Before task scheduling, the IoT devices generate a series of inference requests with different DNN model types $\operatorname{m_{t}^{i}}$, input types $\operatorname{d_{t}^{i}}$ (i.e., image or text), input shapes $\operatorname{d_{s}^{i}}$, and service level objectives $\operatorname{SLO_i}$. 
The $\operatorname{i}$-th request $r_i$, therefore, could be denoted as $\operatorname{r_i=\{m_{t}^{i},d_{t}^{i},d_{s}^{i},SLO_i\}}$. 
Note that requests arrive at BCEdge online at random with a Poisson distribution. 
BCEdge maintains a request queue for each model, and support dynamic batching by aggregating multiple inference requests with the same model into corresponding request queue $\operatorname{seq_b=\{r_1,r_2,…,r_b\}}$, where $\operatorname{b}$ is the batch size of DNN models. 
Meanwhile, BCEdge constructs multiple instances $\operatorname{m_c(c=1,2,…,n)}$ for each model (i.e., concurrent instances), which is critical for edge platforms with GPU, since batching and concurrent inference can effectively improve the throughput. 
The model zoo in BCEdge backend executes DNN inference from the request queue, and returns the prediction result $\operatorname{O_b=\{o_1,o_2,…,o_b\}}$.

\subsubsection{Scheduling Model}
Since the SLO is different for each request, a fixed scheduling time slot is not suitable.
On the other hand, it is not feasible to specify scheduling time slots for each request individually, which significantly increases system overhead.
Therefore, we set the $\operatorname{i}$-th scheduling time slot $\operatorname{t_i}$ as the ratio of the sum of $\operatorname{SLO_i}$ for a batch requests to the number of concurrent models, which denoted as
\begin{equation}
\operatorname{t_i=\sum_{i=1}^{b}SLO_i/m_c}
\label{E:time}
\end{equation}

In this way, BCEdge is capable of guaranteeing the SLO of each request and provide efficient inference services with batching and concurrent inference.
Moreover, BCEdge starts the next scheduling immediately after finishing the current scheduling to reduce the GPU idle.

\subsubsection{Computing and Networking Model}
The end-to-end latency involves the communication between IoT devices and edge platforms, as well as model inference time, which consists of the following components: 
\begin{itemize}
\item
\textbf{\emph{request transmission time}} $\operatorname{t_{t}^{i}}$: the time that IoT devices send the $\operatorname{i}$-th inference request (e.g., image or text, etc.) to edge platforms through the network, which depends on communication bandwidth and the size of input data. 
\item 
\textbf{\emph{request serialization time}}  $\operatorname{t_{s}^{i}}$: the time to aggregate multiple inference requests with the same model into a single request queue at the edge platform, for batching and concurrent inference of model instances.
\item
\textbf{\emph{request queuing time}} $\operatorname{t_{w}^{i}}$: the time that the request is blocked on the request queue until it is scheduled, which relate to batch size and the number of concurrent models.
\item
\textbf{\emph{DNN inference time}} $\operatorname{t_{m}^{i}}$: the time that edge platform execute model inference. Once inference is complete, the current request is removed from the queue.
\item
\textbf{\emph{result transmission time}} $\operatorname{t_{o}^{i}}$: the time that edge platform sends the $\operatorname{i}$-th inference request to IoT devices through the network, which is related to the network bandwidth, regardless of the result size (usually negligible).
\end{itemize}

Thus, the overall latency $\operatorname{t_{r}^{i}}$ can be denoted as:
\begin{equation}
\operatorname{t_{r}^{i}=t_{t}^{i}+t_{s}^{i}+t_{w}^{i}+t_{m}^{i}+t_{o}^{i}}
\label{E:time}
\end{equation}

\subsection{Problem Formulation} \label{problem}
Our objective is to co-optimize both throughput and latency for each DNN model by automatically exploring the feasible set of batch size and number of concurrent models, while guaranteeing SLO.
Inspired by the co-adaptive scheduler named Pollux~\cite{qiao2021pollux}, we present a \emph{utility function} $\operatorname{U}$ in Eq.~(\ref{E:Utility}) to evaluate the trade-off between throughput and latency.
\begin{equation}
\operatorname{U} = \operatorname{log(T_{t_i}(b, m_c)/\frac{L_{t_i}(b, m_c)}{(\sum_{j=1}^{b}r_j)/m_c})}
\label{E:Utility}
\end{equation}
where $\operatorname{b}$ is the batch size, and $\operatorname{m_c}$ is the number of concurrent models. 
The throughput in the $\operatorname{i}$-th scheduling time slot $\operatorname{t_i}$ can be denoted as $\operatorname{T_{t_i}(b, m_c)}$, and $\operatorname{L_{t_i}(b, m_c)}$ represents the actual latency of the $\operatorname{i}$-th request. $\operatorname{{(\sum_{j=1}^{b}r_j)/m_c}}$ denotes the ratio of the sum of SLOs for batch requests to the number of concurrent models. Notably, $\operatorname{\frac{L_{t_i}(b, m_c)}{(\sum_{j=1}^{b}r_j)/m_c}\in(0,1]}$ avoids request scheduling failure as much as possible while ensuring real-time performance.

The scheduler must consider the memory capacity of edge platforms $\operatorname{M_i}$ and the SLO constraints $\operatorname{SLO_i}$, when batching and concurrent executing requests.
Therefore, the optimization objective with above requirements is formulated as
\begin{equation}
\begin{split}
\operatorname{min.} & \operatorname{U} \\
\operatorname{s.t. m_i} & \leq \operatorname{M_i} \\
\operatorname{L_i} & \leq \operatorname{SLO_i}
\label{1}
\end{split}
\end{equation}
where $\operatorname{m_i}$ is the actual used memory for the $\operatorname{i}$-th request, and $\operatorname{L_i}$ is the end-to-end latency of the $\operatorname{i}$-th request.

Table~\ref{Symbol} provides mainly symbol definitions and corresponding descriptions.
\begin{table}
\setlength{\abovecaptionskip}{0pt}
\setlength{\belowcaptionskip}{0pt}
    \caption{Symbol Table and Description}
    \label{Symbol}
    \centering
    \begin{tabular}{r|l} \hline
    	Notation & Description \\ \hline
    	$\operatorname{m_{t}^{i}}$ & DNN model type of the $\operatorname{i}$-th request  \\
    	$\operatorname{d_{t}^{i}}$ & requested input type of the $\operatorname{i}$-th request \\
    	$\operatorname{d_{s}^{i}}$ & requested input shape of the $\operatorname{i}$-th request \\ 
    	$\operatorname{SLO_i}$ & service level objective of the $\operatorname{i}$-th request \\ 
        $\operatorname{seq_b}$ & request queue with batch size $\operatorname{b}$ \\
        $\operatorname{m_c}$ & number of concurrent models \\
        $\operatorname{O_b}$ & inference result \\
        $\operatorname{t_i}$ & the $\operatorname{i}$-th scheduling time slot \\ 
        $\operatorname{t_{r}^{i}}$ & end-to-end latency of the $\operatorname{i}$-th request \\ 
        $\operatorname{u}$ & average resource utilization \\
        $\operatorname{s_i}$ & batching slot of the $\operatorname{i}$-th queue\\
        $\operatorname{T_{t_i}(b, m_c)}$ & throughput \\ 
        $\operatorname{L_{t_i}(b, m_c)}$ & end-to-end latency \\ 
        $\operatorname{U}$ & utility function \\ \hline
    \end{tabular}
    \vspace{-0.1in}
\end{table}

\section{BCEdge Design}\label{Design}
\subsection{System Overiew}
The goal of BCEdge is to devise a scheduling framework for multi-model DNN inference serving, which aims to allocate a moderate batch size and number of concurrent models for each incoming inference requests, while maintaining SLO.
To this end, the scheduling of DNN inference requests with SLO requirements must consider two aspects: batching, and concurrent model instances.
Unlike the prior work which consider a subset of the two dimensions~\cite{yang2021deeprt,crankshaw2017clipper}, we propose a scheduler that fully explores all two dimensions to find the most effective point for scheduling.

\begin{figure}[htbp]
\large
\centerline{\includegraphics[width=\linewidth]{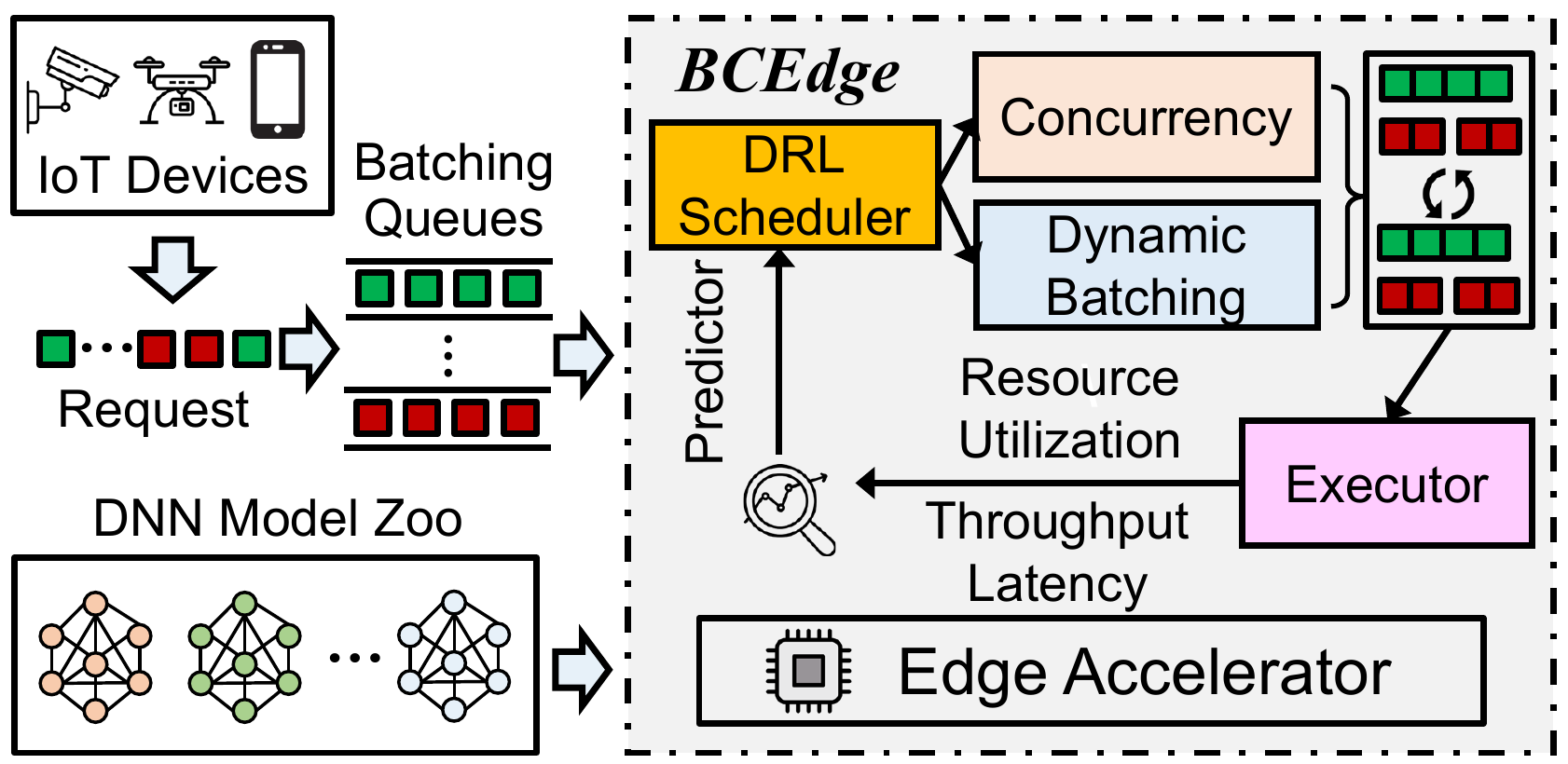}}
\caption{Overview of the scheduling framework for DNN service.}
\label{fig:system}
\vspace{-0.2cm}
\end{figure}

Fig.~\ref{fig:system} presents the overall architecture of our proposed scheduling framework, namely BCEdge.
The framework is composed of learning-based scheduler (Section~\ref{scheduler}), dynamic batching module (Section~\ref{batching}), concurrent instance module (Section~\ref{concurrency}), performance analyzer (Section~\ref{Profiler}), and SLO-aware interference predictor (Section~\ref{Predictor}).
BCEdge first \ding{182} maintain a request queue for each DNN model.
The requests with different DNN models generated by IoT devices are merged to send the corresponding request queue.
The performance profiler \ding{183} periodically collects the information (e.g., utilization, SLO, system throughput and end-to-end latency for a pair of batch size and number of concurrent models) for each DNN model.
Meanwhile, the SLO-aware interference predictor \ding{184} analyzes the potential interference overhead caused by concurrent model instances, which guides the scheduler to make more robust decisions.
The learning-based scheduler then \ding{185} finds the best batch size and number of concurrent models by leveraging profiled information, and feeds back to dynamic batch processing module and concurrent instance module, respectively.
The executor in the backend finally \ding{186} executes DNN inference service with batching and number of concurrent models on the edge platform.

\subsection{Learning-based Scheduler}\label{scheduler}
\textbf{\emph{Search Space Challenge}}:
The learning-based scheduler is the critical component for BCEdge. 
Note that the scheduling in BCEdge is more complex compared with prior work (e.g., 
TF-Serving~\cite{olston2017tensorflow}, Clipper~\cite{crankshaw2017clipper}, and DeepRT~\cite{yang2021deeprt}), since it involves batching as well as concurrent inference.
The challenge of the two-dimensional scheduling space (batch size, and number of concurrent models) for BCEdge is that the scheduling decision is affected by several variables dependent on each other.
Specifically, the best batch size and number of concurrent models depends on the computing requirements of different DNN models, the properties of input, the SLO constraints, as well as the available computing resources of edge platforms.
Therefore, the optimal trade-off configuration would sit on the sweet spot in the search space built upon the two dimensions, which creates a huge search space.

To this end, we tailor the learning-based scheduling algorithm for BCEdge. 
Compared with traditional heuristic methods, deep reinforcement learning (DRL) has great advantages in processing complex policy decision, which can be applied to action spaces with high-dimensional. 
Thus, we design a novel DRL-based scheduler for efficient DNN inference service with batching and concurrent inference, in order to trade off throughput and latency. 
Since batch size and number of concurrent models are discrete, we present a learnable online scheduling algorithm with maximum entropy, based on discrete soft actor-critic~\cite{christodoulou2019soft} framework, which maximizes the reward while maximizing the entropy of the visited states compared with traditional DRL approaches. 
The introduction of entropy makes our proposed scheduling algorithm have the following benefits: 
\begin{itemize}
\item
Enable the agent in DRL to learn more near-optimal actions to accelerate training (i.e., the output is a policy distribution), compared with deterministic policy-based DRL (i.e., the output is an action).
\item 
Enable the agent to has a stronger ability to explore the environment, and avoid falling into local optimum.
\item
Enable the system to be more robust.
\end{itemize}

Now we focus on how the scheduler in BCEdge finds batch size and number of concurrent models for each inference request that optimizes the trade-off in Eq.~(\ref{E:Utility}).
We describe the details as follows:

\subsubsection{Markov Decision Process Formulation}
Firstly, we model batching and concurrent scheduling of inference requests as a markov decision process (MDP). It can be denoted as a five-tuple ($\cal S$, $\cal A$, $\pi$, $p$, $r$):
\begin{itemize}
\item 
\textbf{\emph{State}}: $\cal S$ is the discrete state space. At each scheduling time slot $t_i$, the agent in DRL constructs a state $s_t(s_t\in\cal S)$ to periodically collect request information and the resource utilization of edge platforms. $s_t$ consists of five parts:
(I) The DNN model type $m_{t}^{i}$. 
(II) The input type $d_{t}^{i}$ and input shape $d_{s}^{i}$. 
(III) The SLO of each requests. 
(IV) The available computing resources of edge platforms $m_i$. 
(V) The information of request queue $seq_b$.

\item
\textbf{\emph{Action}}: $\cal A$ is the discrete action space.
The action of the agent in DRL is to find best batch size $b$ and number of concurrent models $m_c$.
The action $a_t(a_t \in \cal A)$ at scheduling time slot $t_i$ can be denoted as $a_t=(b, m_c)$.
For instance, if a DNN model has $\cal M$ optional batch sizes and $\cal N$ optional number of concurrent models, the size of the discrete action space $\cal A$ is $\cal M\times\cal N$.

\item
\textbf{\emph{Policy}}: The policy $\pi\left(a_{t} \mid s_{t}\right)$ is a function that the agent decides the next action $a_t$ according to the environment state $s_t$ at timestamp \emph{t}.
In the maximum entropy-based DRL algorithm, we maximize both the reward and the entropy of the visited states. The optimal policy $\pi^*$ is denoted as follows:
\begin{equation}
\begin{split}
\pi^{*}= \operatorname{argmax}_{\pi} \sum_{t=0}^{T} E_{(s_{t}, a_{t}) \sim \rho_{\pi}}[\gamma^{t} r(s_{t}, a_{t}) \\ + \alpha \mathcal{H}(\pi(\cdot \mid s_{t}))]
\end{split}
\label{Policy}
\end{equation}
where $\gamma\in[0,1]$ is discount factor, $\rho_\pi$ represents the distribution of action trajectory generated by the policy $\pi$, and $\alpha$ is a temperature parameter to express
the relative importance of reward and entropy.
$\mathcal{H}\left(\pi\left(\cdot \mid s_{t}\right)\right)=-\log \pi\left(\cdot \mid s_{t}\right)$, which is the entropy with state $s_t$.

\item
\textbf{\emph{State transition probability}}: $p\left(s_{t}^{\prime} \mid s_{t}, a_{t}\right)$ is the state transition probability that indicates the probability of transitioning to the next state $s_t^{\prime}$ after taking an action $a_t$ in the current state $s_t$ at timestamp \emph{t}, satisfying $\sum_{s^{\prime} \in S} p\left(\left.s_{t}^{\prime}\right|s_{t}, a_{t}\right)=1$.

\item
\textbf{\emph{Reward}}: $r: \mathcal{S} \times \mathcal{A} \rightarrow \mathbb{R}$ is the reward function.
The agent in DRL aims to maximize the accumulated expected reward $\mathbb{E}\left[\sum_{t=0}^{T} \gamma^{t} r_{t}\right]$, where $r_t$ is the instant reward when the agent selects batch size and number of concurrent models at each scheduling time slot $t_i$. 
Since our objective is to maximize the trade-off between throughput and latency, we migrate the objective in Eq.~(\ref{E:Utility}) to the reward function:
\begin{equation}
r_{t}=U
\label{Reward}
\end{equation}
\end{itemize}

In this way, the traditional scheduling problem is converted into the maximization of reward in DRL. 
We further utilize efficient learning-based algorithm to reduce the complexity.

\subsubsection{Maximum Entropy DRL-based scheduling algorithm}
We leverage soft policy iteration~\cite{haarnoja2018soft} to maximize both reward and entropy.
To be more specific, soft policy iteration consists of policy evaluation and policy improvement, which alternates during training.
\begin{itemize}
\item 
\textbf{\emph{Soft Q-Function (Critic Network)}}: We first compute the soft Q-value $Q(s_t, a_t)$ in the policy evaluation step. The soft Q-function is denoted as follows~\cite{haarnoja2018soft}:
\begin{equation}
\mathcal{T}^{\pi} Q\left(\mathbf{s}_{t}, \mathbf{a}_{t}\right) \triangleq r\left(\mathbf{s}_{t}, \mathbf{a}_{t}\right)+\gamma \mathbb{E}_{\mathbf{s}_{t+1} \sim p}\left[V\left(\mathbf{s}_{t+1}\right)\right]
\label{soft q function}
\end{equation}
where $\mathcal{T}^{\pi}$ is the modified bellman backup operator.

The soft state value function $V(s_t)$ with the policy $\pi$ in discrete state space $s_t$ is:
\begin{equation}
V\left(s_{t}\right):=\pi\left(s_{t}\right)^{T}\left[Q\left(s_{t}\right)-\alpha \log \left(\pi\left(s_{t}\right)\right)\right]
\label{State Value function}
\end{equation}

We train the soft Q-value in Eq.~(\ref{soft q function}) by minimizing the soft bellman residual, and the loss function is: 
\begin{equation}
\begin{split}
J_{Q}(\theta)=E_{(s_{t}, a_{t}) \sim D}[\frac{1}{2}(Q_{\theta}(s_{t}, a_{t}) \\
-(r(s_{t}, a_{t})+\gamma E_{s_{t+1} \sim p(s_{t}, a_{t})}[V_{\bar{\theta}}(s_{t+1})]))^{2}]
\label{gradient_Q}
\end{split}
\end{equation}

\item 
\textbf{\emph{Policy (Actor Network)}}: The policy improvement is used to update the policy network, which is denoted as follows:
\begin{equation}
\pi_{\text {new }}=\arg \min _{\pi^{\prime} \in \Pi} \mathrm{D}_{\mathrm{KL}}\left(\pi^{\prime}\left(\cdot \mid \mathbf{s}_{t}\right) \| \frac{\exp \left(\frac{1}{\alpha} Q^{\pi_{\text {old }}}\left(\mathbf{s}_{t}, \cdot\right)\right)}{Z^{\pi_{\text {old }}}\left(\mathbf{s}_{t}\right)}\right)
\label{Policy_improve}
\end{equation}
where $\mathrm{D}_{\mathrm{KL}}$ is the Kullback-Leible (KL) divergence, and $Z^{\pi_{\text {old }}}$ is the partition function.

We minimize the KL divergence in Eq.~(\ref{Policy_para}) to update the parameters of the policy network:
\begin{equation}
J_{\pi}(\phi)=E_{s_{t} \sim D_{KL}}\left[\pi_{t}\left(s_{t}\right)^{T}\left[\alpha \log \left(\pi_{\phi}\left(s_{t}\right)\right)-Q_{\theta}\left(s_{t}\right)\right]\right]
\label{Policy_para}
\end{equation}

\item 
\textbf{\emph{Temperature parameter}}: We automatically adjust the temperature parameter $\alpha$ in Eq.~(\ref{Policy}), according to~\cite{haarnoja2018soft}:
\begin{equation}
J(\alpha)=E_{a_{t} \sim \pi_{t}}\left[-\alpha\left(\log \pi_{t}\left(a_{t} \mid s_{t}\right)+\bar{H}\right)\right]
\label{Temperature}
\end{equation}
where $\bar{H}$ is a constant vector that equals to the hyperparameters of the target entropy.
\end{itemize}

Algorithm~\ref{alg:alg1} provides the overall procedure of scheduling concurrent model instances with dynamic batching.
The scheduler first receives the information of DNN model and resource utilization for each inference request.
Before each scheduling time slot, it initializes all networks, including soft Q-network, target soft Q-network, policy network and temperature network with corresponding parameters, respectively.
Note that we use two soft Q-networks and take the minimum value of them to alleviate the overestimation of soft Q-value.

For each scheduling time slot, the scheduler first checks each request queue.
If the request queue is empty, it pushes incoming requests into the request queue (\emph{line 7}).
The scheduler then takes an actions (e.g., determine the best batch size and number of concurrent models for each request) based on Eq.~(\ref{Policy}), and the agent in DRL obtains a instant reward as utility (\emph{line 9}).
Meanwhile, the state changes from $s_t$ to $s_{t+1}$, and the current state, action, reward and the next state are stored as a action transition in the replay buffer $\mathcal{D}$.
The scheduler pulls the request sequence from the batching slot (Section~\ref{batching}) when the batch requests in the current request sequence are executed (\emph{line 12}).
The scheduler finally update the parameters of all networks, and repeat the above process (\emph{line 14$\sim$18}) until the end of the iteration.

\begin{algorithm}
\SetKwInOut{Input}{Input}
\SetKwInOut{Output}{Output}
\caption{Learning-based scheduling algorithm}
\label{alg:alg1}
\Input{The information ($m_{t}^{i}$, $d_{t}^{i}$, $d_{s}^{i}$, $SLO_i$) of request $r_i$, resource utilization $u$}
\Output{batch size $b$, number of concurrent models $m_c$}
Initialization all neural networks $Q_{\theta_{1}}, Q_{\theta_{2}}, \hat{Q}_{\theta_{1}}, \hat{Q}_{\theta_{2}}, \pi_{\phi}, T_{\alpha}$\;
Randomly initialize network parameters $\theta_{1}, \theta_{2}, \phi, \alpha$\;
Initialize an empty replay buffer $\mathcal{D} \leftarrow \emptyset$\;
\For{each scheduling time slot $t_i$}{
  \For{each environment step}{
    \If{request queue $seq_b == \emptyset$}{
        Push requests $r_i$ to request queue $seq_b$\;
    }
    Take an action $a_t(b, m_c)$ based on policy $\pi$ and get reward $r_t\left(a_{t} \mid s_{t}\right)$ using Eq.~(\ref{Reward})\;
    $s_{t+1} \sim p\left(s_{t+1} \mid s_{t}, a_{t}\right)$ \;
    $\mathcal{D} \leftarrow \mathcal{D} \cup\left\{\left(s_{t}, a_{t}, r\left(s_{t}, a_{t}\right), s_{t+1}\right)\right\}$\;
    Pull current request queue $seq_b$ from batching slot $s_i$\;
 }
  \For{each gradient step}{
    Update actor and target networks $\theta_{i}$ for $i \in\{1,2\}$ using Eq.~(\ref{gradient_Q})\;
    Update critic network $\phi$ using Eq.~(\ref{Policy_para})\;
    Update temperature network $\alpha$ using Eq.~(\ref{Temperature})\;
  }
}
\end{algorithm}

\subsection{Dynamic Batching} \label{batching}
\begin{figure}[htbp]
\vspace{-0.2cm}
\large
\centerline{\includegraphics[width=0.9\linewidth]{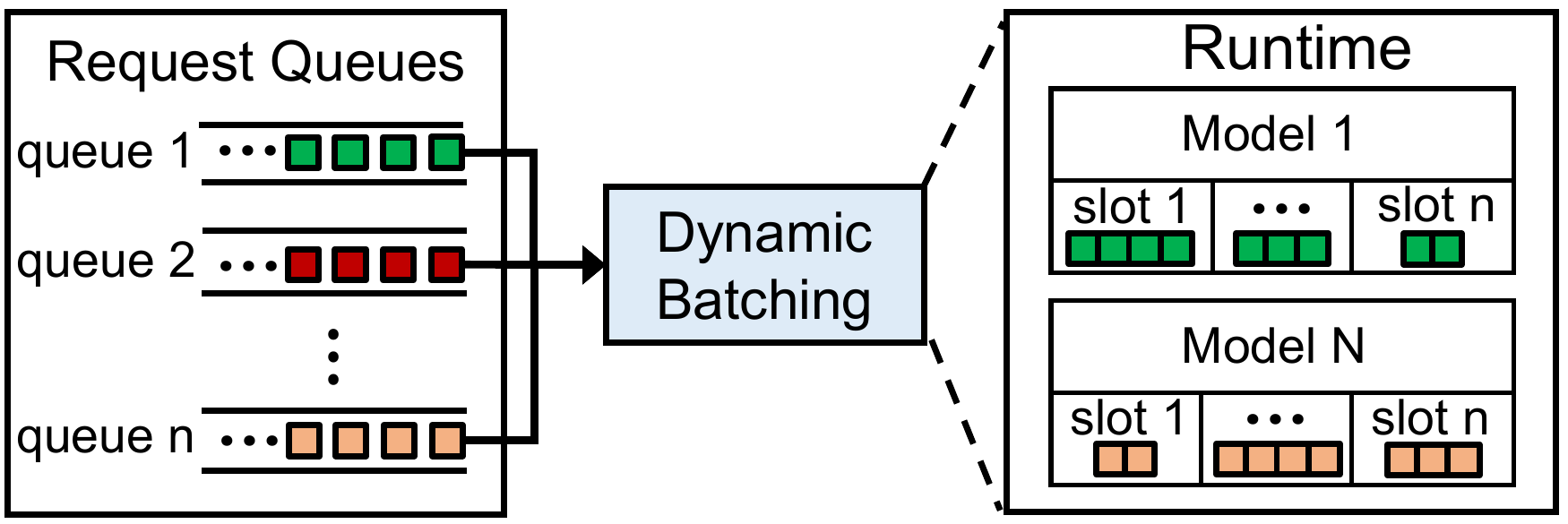}}
\caption{Dynamic batching module.}
\label{fig:batch_model}
\vspace{-0.2cm}
\end{figure}

BCEdge enables batch inferencing by allowing individual inference requests to specify a batch of inputs. The inferencing for a batch of inputs is performed at the same time which is especially important for GPUs, since it can greatly increase inferencing throughput. 
As illustrated in Fig.~\ref{fig:batch_model}, the dynamic batching maintains a request queue separately for requests with different models, and each batch size in a queue depends on the learning-based scheduler in BCEdge.
The dynamically created batches are distributed to all model instances configured for the model, and dynamic batching module then concurrently executes multiple batch request queues for each model. 
To be more specific, dynamic batching first adds each requests to the corresponding request queue based on the order of arrival. 
Meanwhile, it sorts the priority based on the SLO of inference requests in each queue, the shorter the SLO, the higher the priority. 
Dynamic batching then merges multiple requests to a single large request, and assigns batch requests in the request queue to multiple slots of corresponding models at runtime. 
Note that the batch requests are scheduled in the order of arrival if have the same priority.

\subsection{Model Instance Concurrency} \label{concurrency}
\begin{figure}[htbp]
\vspace{-0.2cm}
\centerline{\includegraphics[width=0.6\linewidth]{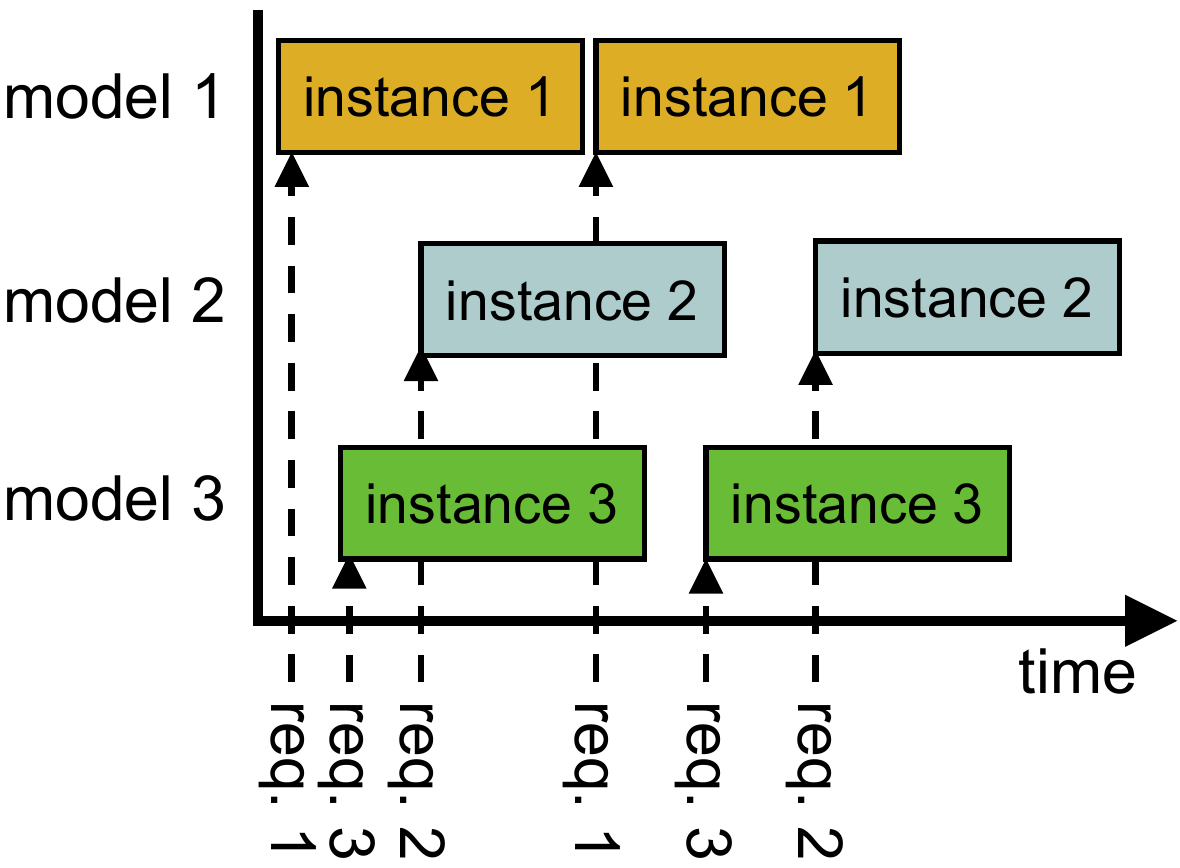}}
\caption{Concurrent instance module.}
\label{fig:concurrent_model}
\vspace{-0.2cm}
\end{figure}

BCEdge enables multiple models and multiple instances of the same model to execution in parallel on single or multiple GPUs, and the DNN models executed on CPU are handled similarly by BCEdge.
Fig.~\ref{fig:concurrent_model} shows the pipeline of executing model instance with batch requests in parallel for three DNN models, and each model is assigned two instances. 
We assume that BCEdge is not currently processing any requests. 
When the first three requests arrive at the same time, each instance of the three models processes a corresponding request. 
BCEdge then immediately dispatches both of them to the GPU, and the hardware scheduler of GPU begins working on three inferences in parallel.
Note that the first three inference requests are immediately executed in parallel, and the last three inference requests must wait until one of the first three requests completes before beginning.
In particular, if multiple inference requests for the same model arrive at the same time, BCEdge serializes their execution by scheduling only one at a time.

\subsection{Performance Profiler} \label{Profiler}
The profiler in BCEdge periodically collects the performance information online, including the current utilization of CPU, GPU, memory, as well as system throughput and end-to-end latency. 
Specifically, the profiler records the performance information of each batch request with different input shapes on the edge platform, and feeds the information back to the scheduler.
The scheduler learns the above information to schedule the next batch request, i.e., determining the best batch size and number of concurrent models to maximize the utility. 
Meanwhile, BCEdge can avoid system overload and improve resource utilization by the performance profiler, which reflects the potential advantage of BCEdge for dynamic resource management and allocation.

\subsection{SLO-Aware Interference Predictor} \label{Predictor}
Concurrent inference of multiple models or multiple instances of a single model can process more requests simultaneously to improve throughput. However, an important challenge is the interference performance caused by concurrent inference of multiple models on a single GPU.
As shown in Fig.~\ref{observation}, we observed that concurrent inference significantly increases latency compared to executing a single model independently, as multiple models compete for the shared resources on edge platform, especially the memory. 
In such case, model interference may cause the scheduler to make incorrect schedules, and may violate the SLO.

\begin{figure}[htbp]
\large
\centerline{\includegraphics[width=0.8\linewidth]{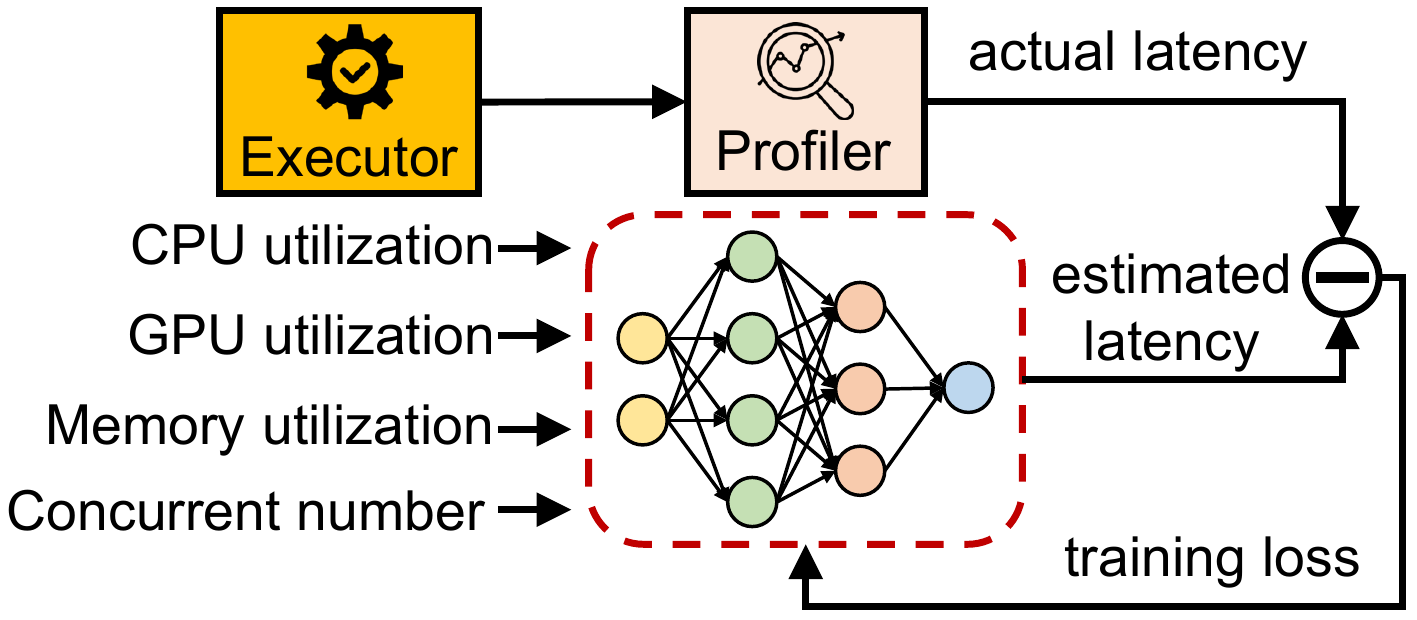}}
\caption{SLO-aware interference predictor based on neural network.}
\label{fig:predictor}
\vspace{-0.2cm}
\end{figure}

A key challenge in mitigating interference is to predict latency increases when multiple inferences are executed concurrently in the same GPU. 
To confine the interference effect, we utilize a lightweight two-layer neural network (NN) with negligible overhead as the predictive model, which directly learns the interference latency of concurrently executing multiple inferences on a single GPU.
As shown in Fig.~\ref{fig:predictor}, the simple yet effective interference-prediction model based on NN utilizes the currently available computing resources (i.e., memory, CPU and GPU) and the number of concurrent models learned by scheduler as the input of the neural network.
We then compare the estimated latency of the neural network output with the actual latency based on performance feedback provided by the performance profiler, and the neural network is trained by minimizing the standard deviation between the real values and the estimation value.
The trained neural network aims to improve the stability of the scheduler and reduce the SLO violation rate.

\section{Evaluation}\label{Evaluation}
\subsection{Experiment setup}
\begin{figure}[htbp]
\large
\centerline{\includegraphics[width=0.8\linewidth]{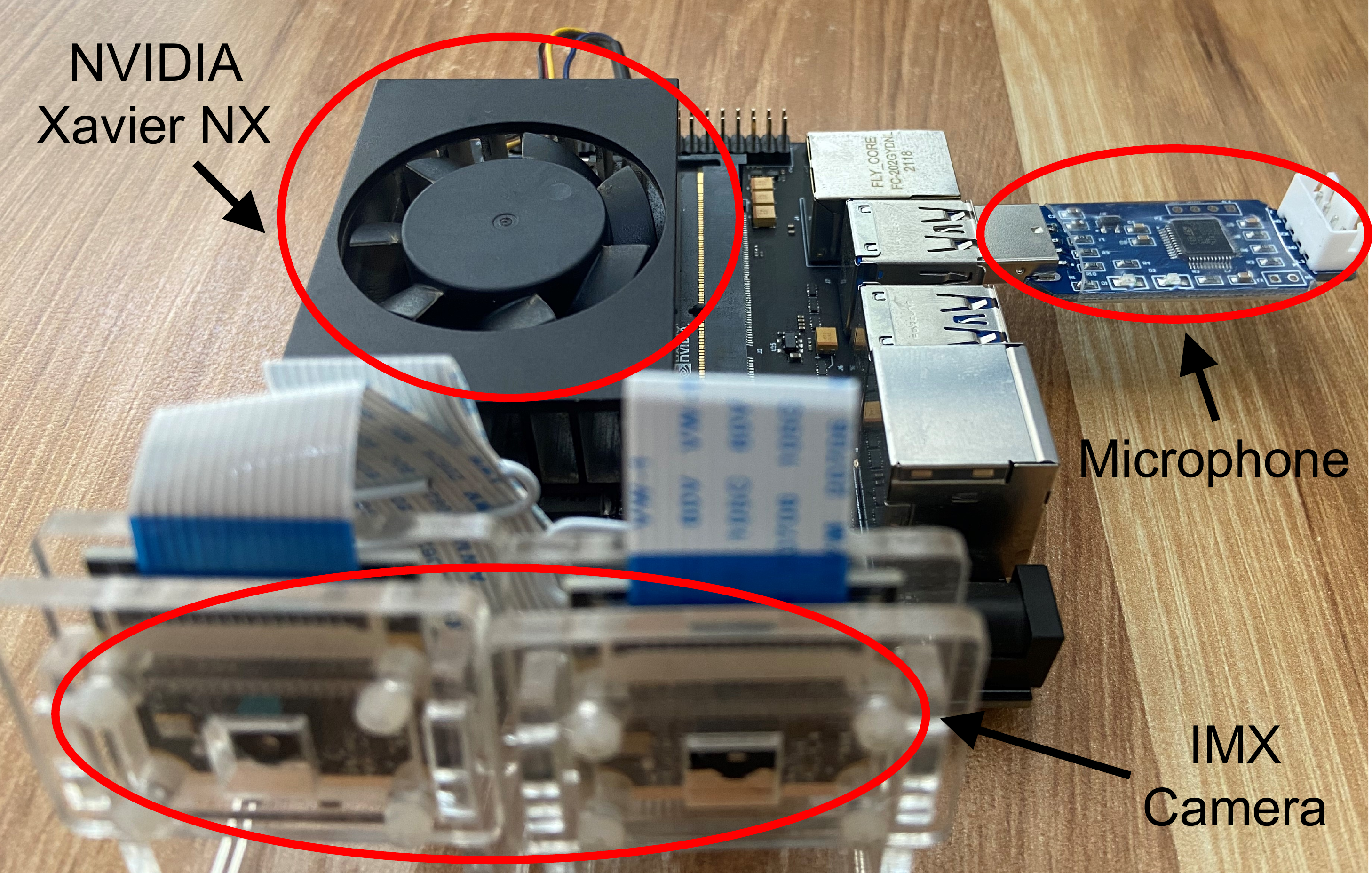}}
\caption{BCEdge prototype implemented on NVIDIA Xavier NX edge platform. We use two IMX cameras and a micorphone as IoT devices.}
\label{fig:prototype}
\end{figure}

\textbf{\emph{BCEdge Prototype:}} We implement the prototype of BCEdge as a runtime backend for Triton~\cite{Triton}, a inference serving system from NVIDIA.
Table~\ref{Experiment} provides a detailed description of the evaluated inference system and the used GPU specification.
The table also provides the versions of the operating system, CUDA, runtime, and machine learning framework.
As Fig.~\ref{fig:prototype} shows, we use two IMX cameras and a micorphone as IoT devices.
The request arrival rate is set to 30 requests per second (rps), and follows the Poisson random distribution. 
Unless otherwise indicated, all evaluations are reported on a NVIDIA Xavier NX edge GPU. 

\textbf{\emph{DNN Models:}} We use six DNN models for three popular DNN families to process image and speech data.
Specifically, we use YOLO-v5~\cite{YOLOv5} for object detection task, MobileNet-v3~\cite{howard2017mobilenets}, ResNet-18~\cite{he2016deep}, EfficientNet-B0~\cite{tan2019efficientnet} and Inception-v3~\cite{szegedy2016rethinking} for image classification tasks, especially with TinyBERT~\cite{jiao-etal-2020-tinybert} for speech recognition tasks.
We use TensorRT~\cite{TensorRT} to reduce memory footprint for better batching and concurrent executing.
Since the limited computing power of Xavier NX, we downsample the input image to 224$\times$224 resolution.
All images are colored frames with the 3 RGB channels. 
Each corresponding SLO latency is listed in Table~\ref{model}.

\textbf{\emph{Training Details:}} Our proposed Algorithm~\ref{alg:alg1} is based on the SAC~\cite{christodoulou2019soft} framework.
All networks are trained using the Adam optimizer with a learning rate of $10^{-3}$.
Each network has a two-layer ReLU neural network with 128 and 64 hidden units, respectively, and the buffer size is fixed to $10^{6}$.
We trained it offline on an off-the-edge device using four NVIDIA GeForce GTX 3080 GPUs with a mini-batch size of 512 for 500 epochs.
We then deploy trained algorithm online to edge platform.
\begin{table}
\setlength{\abovecaptionskip}{0pt}
\setlength{\belowcaptionskip}{0pt}
    \caption{The evaluated system specifications}
    \label{Experiment}
    \centering
    \begin{tabular}{rl} \hline
    	\textbf{Edge Platform} & NVIDIA Xavier NX \\ \hline
    	\textbf{Operating System} & Ubuntu: 18.04.6 (kernel 4.15.0)  \\
    	\textbf{Software} & CUDA 10.2 and TensorRT 8.2~\cite{TensorRT}   \\
        \textbf{CPU} & 6-core Carmel ARMv8.2 \\
        \textbf{GPU} & 384-core Volta GPU with 48 Tensor Cores \\
    	\textbf{Memory Capacity} & 8GB RAM  \\
        \textbf{Runtime} & NVIDIA Triton Inference Server 2.19.0~\cite{Triton}  \\
    	\textbf{ML Framework} & PyTorch 1.10 \\  \hline
    \end{tabular}
    \vspace{-0.1in}
\end{table}

\begin{table}
\setlength{\abovecaptionskip}{0pt}
\setlength{\belowcaptionskip}{0pt}
    \caption{List of DNN models used in the evaluation}
    \label{model}
    \centering
    \begin{tabular}{rlc} \hline
    	\textbf{Model}         & \textbf{Input Shape (Dimension)} & \textbf{SLO (ms)} \\ \hline
    	YOLO-v5 (yolo)         & VOC-2012 (3x224x224)             & 138 \\
    	MobileNet-v3 (mob)     & ImageNet-2012 (3x224x224)        & 86  \\ 
        ResNet-18 (res)        & ImageNet-2012 (3x224x224)        & 58  \\  
        EfficientNet-B0 (eff)  & ImageNet-2012 (3x224x224)        & 93  \\ 
        Inception-v3 (inc)     & ImageNet-2012 (3x224x224)        & 66  \\
        TinyBERT (bert)        & Speech Commands (1x14)           & 114 \\ \hline
    \end{tabular}
    \vspace{-0.1in}
\end{table}

\subsection{Baselines}
\subsubsection{\textbf{Edge inference service framework}}
We compare BCEdge with two SOTA edge inference service frameworks:
\begin{itemize}
\item 
\textbf{\emph{DeepRT}}~\cite{yang2021deeprt}: A soft real-time scheduler that adopts dynamic batching with earliest-deadline-first (EDF) scheduling algorithm to execute batch requests.
\item 
\textbf{\emph{Triton with Actor-Critic (TAC)}}~\cite{Triton}: Since Triton only supports manually setting a fixed batch size and number of concurrent models, we combine Triton with Actor-Critic without entropy to compare with BCEdge.
\end{itemize}

\subsubsection{\textbf{Various scheduling algorithms}}
We ported the traditional heuristics and other reinforcement learning methods in BCEdge inference service framework to compare with our scheduling algorithm, including:
\begin{itemize}
\item 
\textbf{\emph{Genetic Algorithm (GA)}}~\cite{whitley1994genetic}: As a search algorithm for optimization problems, the main idea of GA is "survival of the fittest" in the theory of biological evolution. 
We take the fitness function in GA as our proposed utility.
\item 
\textbf{\emph{Proximal Policy Optimization (PPO)}}~\cite{schulman2017proximal}: PPO is an on-policy (the optimization policy and behavior policy of agent in the learning process are the same policy) DRL algorithm based on the Actor-Critic architecture.
\item 
\textbf{\emph{Double Deep Q Network (DDQN)}}~\cite{van2016deep}: As an off-policy (the optimization policy and behavior policy of agent in the learning process are different policies) DRL algorithm, DDQN eliminates overestimation by decoupling the selection of actions in target Q-value and the calculation of target Q-value.
\end{itemize}

\subsection{Trade-off Performance}
\subsubsection{\textbf{Comparison of Edge Inference Frameworks}}
\begin{figure}[htbp]
\centerline{\includegraphics[width=\linewidth]{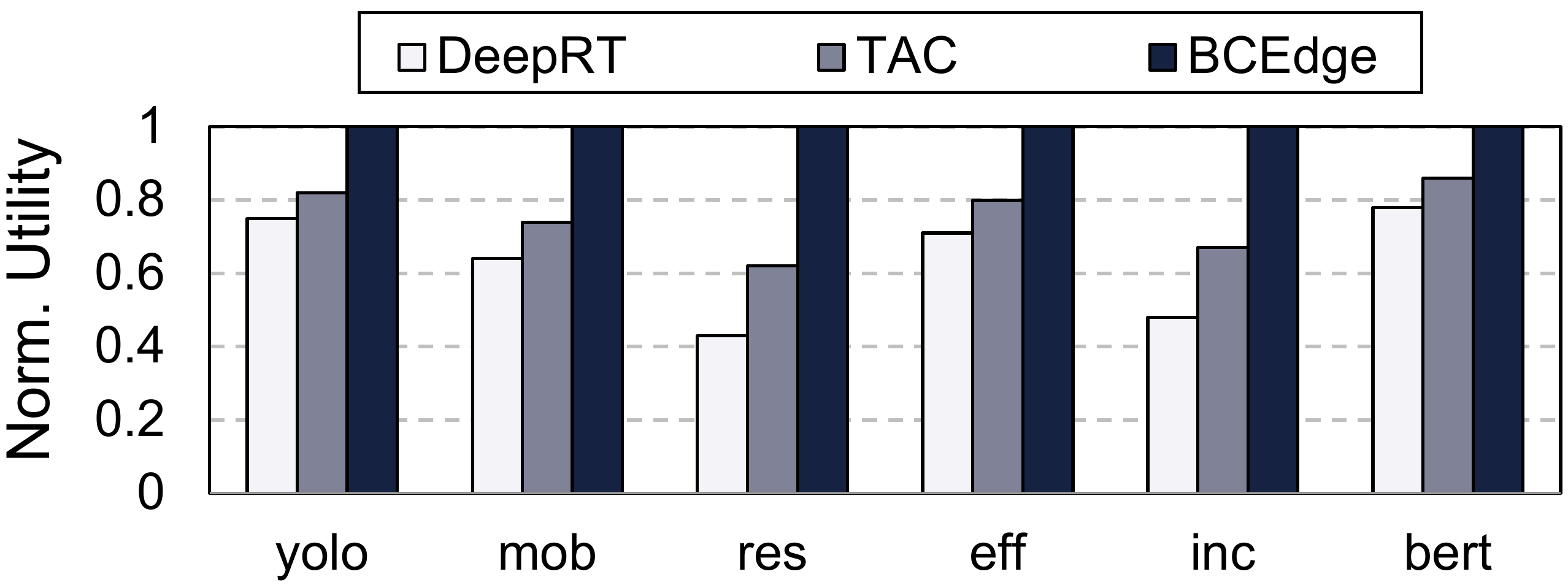}}
\caption{Comparison of the normalized utility with six DNN models.}
\label{fig:sota}
\end{figure}

We first evaluate the performance of BCEdge in terms of the tradeoff between throughput and latency.
Fig.~\ref{fig:sota} reports the normalized utility for six DNN models in Table~\ref{model}. 
Our proposed BCEdge consistently outperforms TAC and DeepRT for all models. 
The lower-utility of DeepRT is caused by the lack of concurrent inference. 
Although TAC leverages a learning-based approach for batching and concurrent scheduling, its agent lacks the entropy that comprehensive explore the environment. 
In contrast, BCEdge introduces entropy into our learning-based scheduling algorithm, so that the agent in DRL have stronger exploration to obtain higher utility. 
In contrast, BCEdge provides a better trade-off between throughput and latency by efficient scheduling as well as SLO-aware interference prediction.
To be more specific, BCEdge offers higher utility than both DeepRT and TAC by an average of 37\% and 25\%, respectively. 

\begin{figure}[htbp]
\centerline{\includegraphics[width=0.85\linewidth]{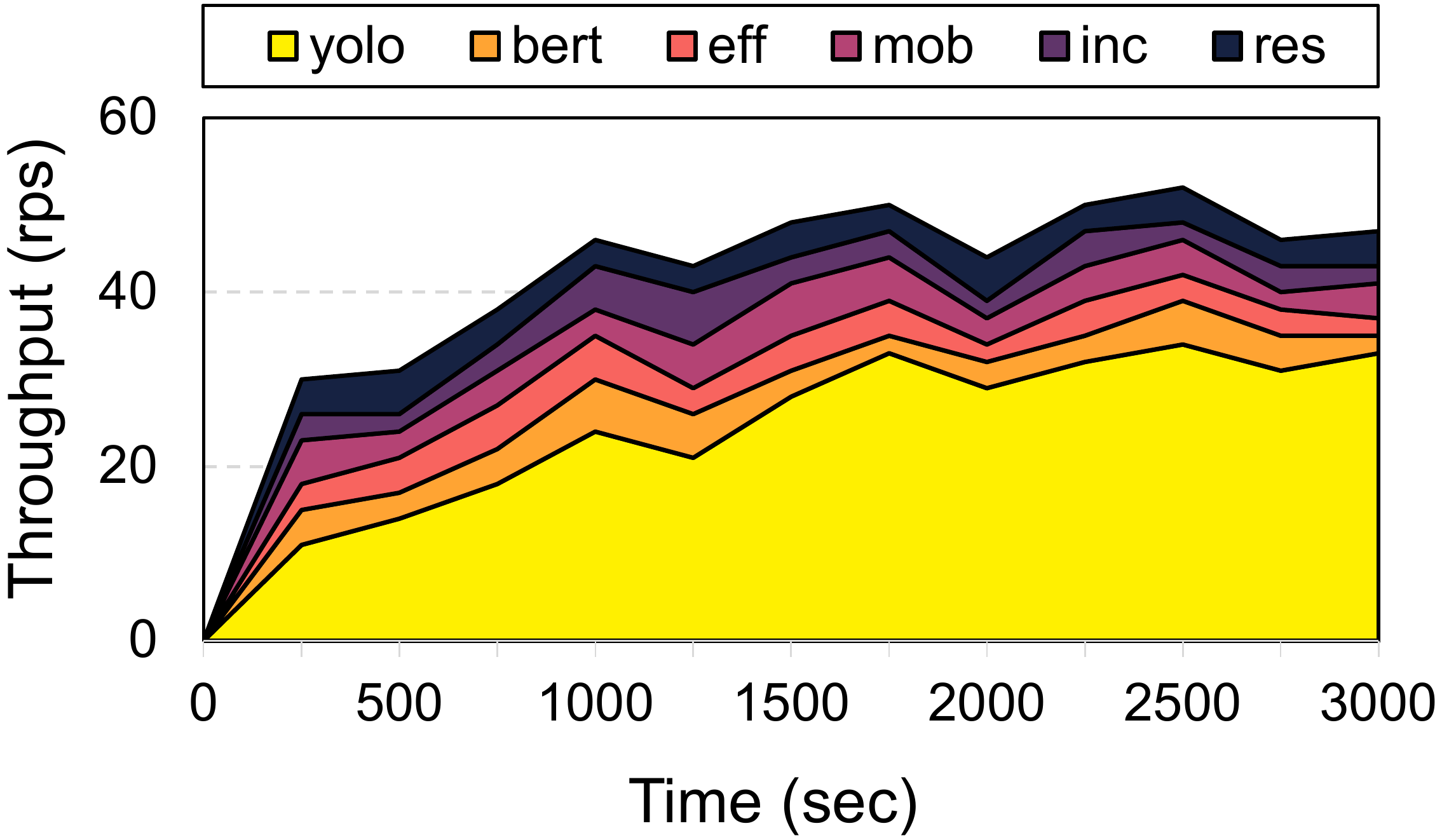}}
\caption{Comparison of throughput with six DNN models. The scheduling duration of each model for 3,000 seconds.}
\label{fig:throughput}
\end{figure}

\begin{figure}[htbp]
\centerline{\includegraphics[width=0.8\linewidth]{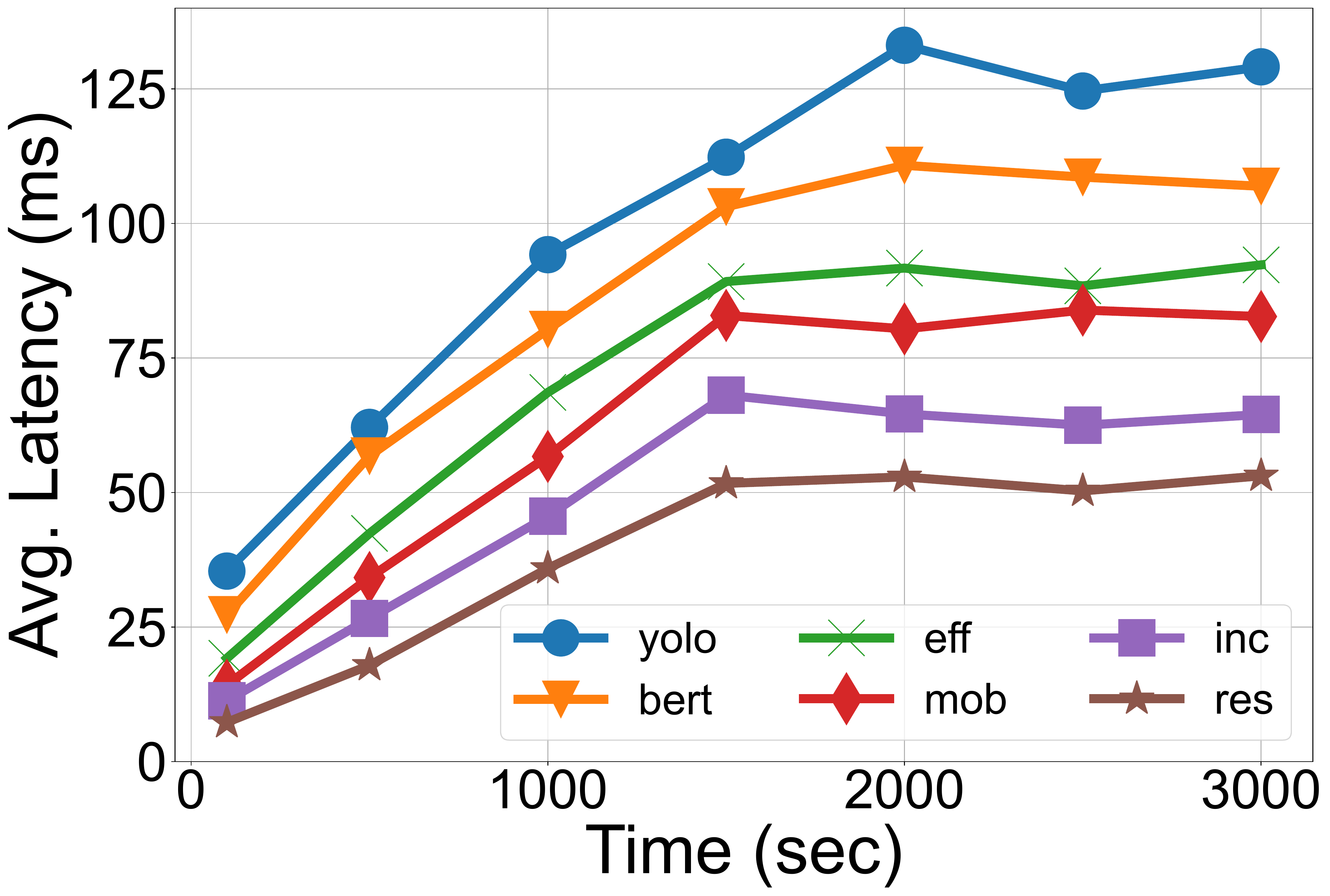}}
\caption{Comparison of average latency with six DNN models. The scheduling duration of each model for 3,000 seconds.}
\label{fig:latency}
\vspace{-0.2cm}
\end{figure}

We next demonstrate that how BCEdge performed six DNN models for a duration of 3,000 seconds in terms of throughput and latency, respectively.
Fig.~\ref{fig:throughput} shows a stacked graph of the accumulated throughput of each model,
and Fig.~\ref{fig:latency} reports the end-to-end latency of each model over time.
Both the throughput and latency increase asymptotically between 0 and 1,500 seconds, which indicates that BCEdge is continuously optimizing our proposed utility function to find the appropriate batch size and number of concurrent models for each DNN model.
Starting from 1,500 seconds, both the throughput and latency are saturated, which indicates that BCEdge has successfully fond the best batch size and number of concurrent models within the constraint of resource and SLO.
In addition, we note that BCEdge tends to sacrifice higher-throughput for lower-latency to achieve better utility.

\subsubsection{\textbf{Comparison of Convergence Performance}}
\begin{figure}[htbp]
\centerline{\includegraphics[width=0.7\linewidth]{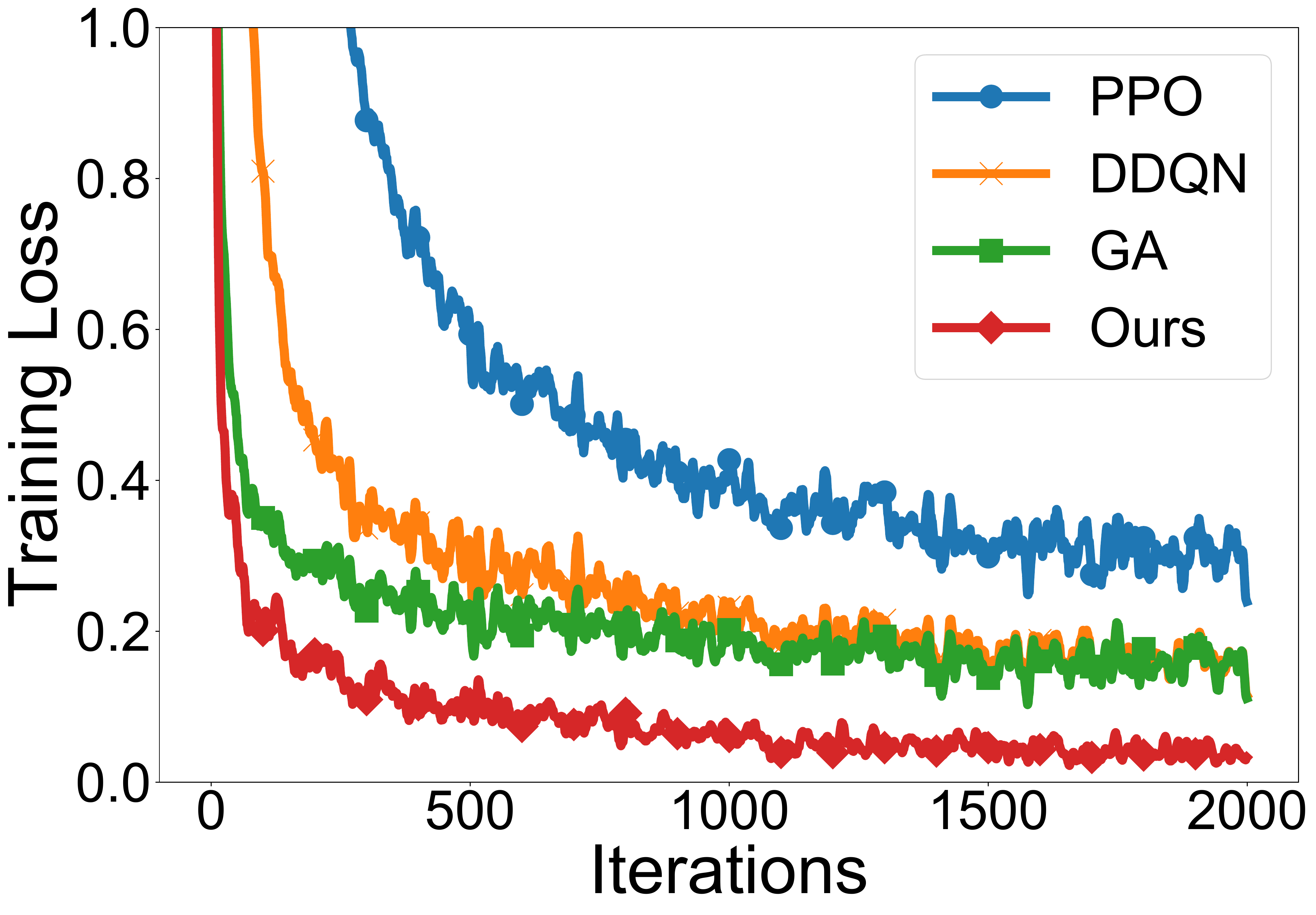}}
\caption{Comparison of training loss for DRL-based (PPO, DDQN and Ours) and heuristics (GA) scheduling algorithms.}
\label{fig:loss}
\vspace{-0.2cm} 
\end{figure}

We compare the convergence of three learning-based (PPO, DDQN and Ours) and heuristic-based (GA) scheduling algorithms in BCEdge DNN inference service framework. 
As shown in Fig.~\ref{fig:loss}, our proposed scheduling algorithm has has the convergence speed increase of 1.8$\times$$\sim$3.7$\times$ compared with baseline methods.
This is due to the entropy enables the agent in DRL to learn more approximate optimal actions. 
That is, there may be multiple actions that are optimal in some states, and our proposed scheduling algorithm makes these actions have the same probability to being selected.
Thus, the maximum entropy can effectively speed up the learning process.
Note that the genetic algorithm (GA) has the disadvantage of being premature, that is, GA has limited ability to explore the environment, therefore it inevitably converge to a local optimal solution.
Importantly, GA involves a large number of calculations, such as crossover, mutation, and etc., resulting in slower convergence.

\subsection{Evaluation of Scalability}
\begin{figure}[htbp]
\centerline{\includegraphics[width=\linewidth]{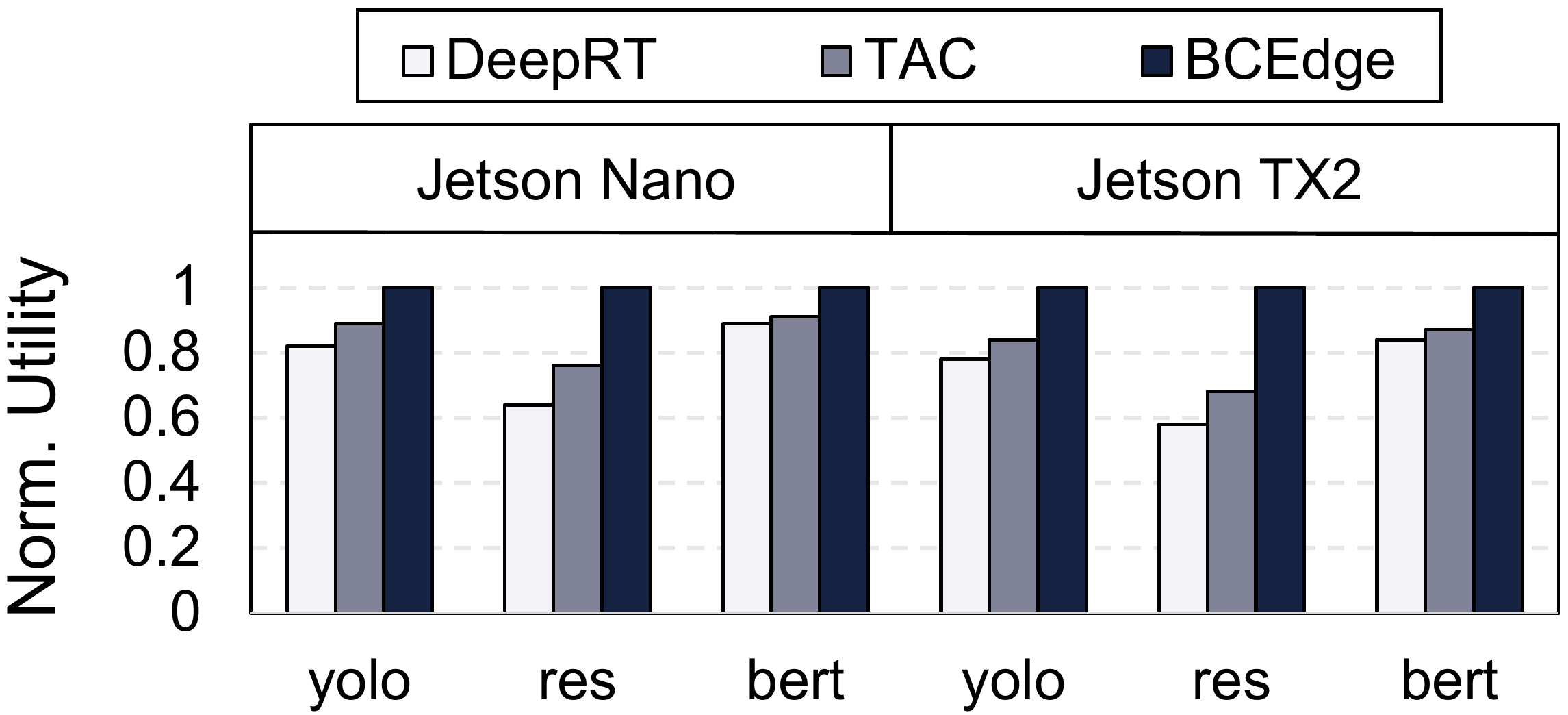}}
\caption{The utility of heterogeneous edge platforms. The more computing resources of the edge platform, the higher the utility.}
\label{fig:device_utility}
\end{figure}

To evaluate scalability beyond our platform in Table~\ref{Experiment}, we additionally select two edge platforms with GPUs (e.g., NVIDIA Jetson Nano and TX2). 
Table~\ref{device} provides the specific parameters of both two heterogeneous edge platforms compared with Xavier NX.
We evaluate the scalability of BCEdge using object detection (YOLO-v5), image classification (ResNet-18) and speech recognition (TinyBERT) DNN models, respectively.
Fig.~\ref{fig:device_utility} reports the utility of BCEdge on Jetson Nano/TX2 edge platforms compared with baselines. 
We can see that BCEdge outperforms the baselines on both two heterogeneous edge platforms. 
Since the image classification DNN model has the least computing resources, that is, ResNet-18 has more batch size and number of concurrent models to be configured than YOLO-v5 and TinyBERT, therefore BCEdge achieves better trade-off for ResNet-18. 
There are similar results in Fig.~\ref{fig:sota}.
Even for Jetson nano with the weakest computing power, the utility of BCEdge can increases by 30\% and 19\% compared with DeepRT and TAC, respectively. 
Since Jetson TX2 has more computing resources that configure more batch size and number of concurrent models, BCEdge can achieve higher performance, and the utility is 39\% and 27\% higher than DeepRT and TAC, respectively.
\begin{table}
\setlength{\abovecaptionskip}{0pt}
\setlength{\belowcaptionskip}{0pt}
\caption{Performance parameters of edge platforms}
\label{device}
\centering
\begin{tabular}{crcc} \hline
\textbf{Edge Platform}  & \textbf{Computility}  & \textbf{Memory}  & \textbf{CUDA Cores} \\ \hline
Jetson Nano             & 0.47 TFLOPS (FP16)    & 4 GB             & 128                 \\
Jetson TX2              & 1.33 TFLOPS (FP16)    & 8 GB             & 256                 \\
Xavier NX               & 21 TOPS (INT8)        & 8 GB             & 384                 \\ \hline
\end{tabular}
\vspace{-0.1in}
\end{table}

\begin{figure} 
\vspace{0pt}
\setlength{\abovecaptionskip}{0pt}
\setlength{\belowcaptionskip}{0pt}
\centering
\subfigure[YOLO-v5 (yolo)]{
\begin{minipage}[b]{\linewidth}
\includegraphics[width=1\linewidth]{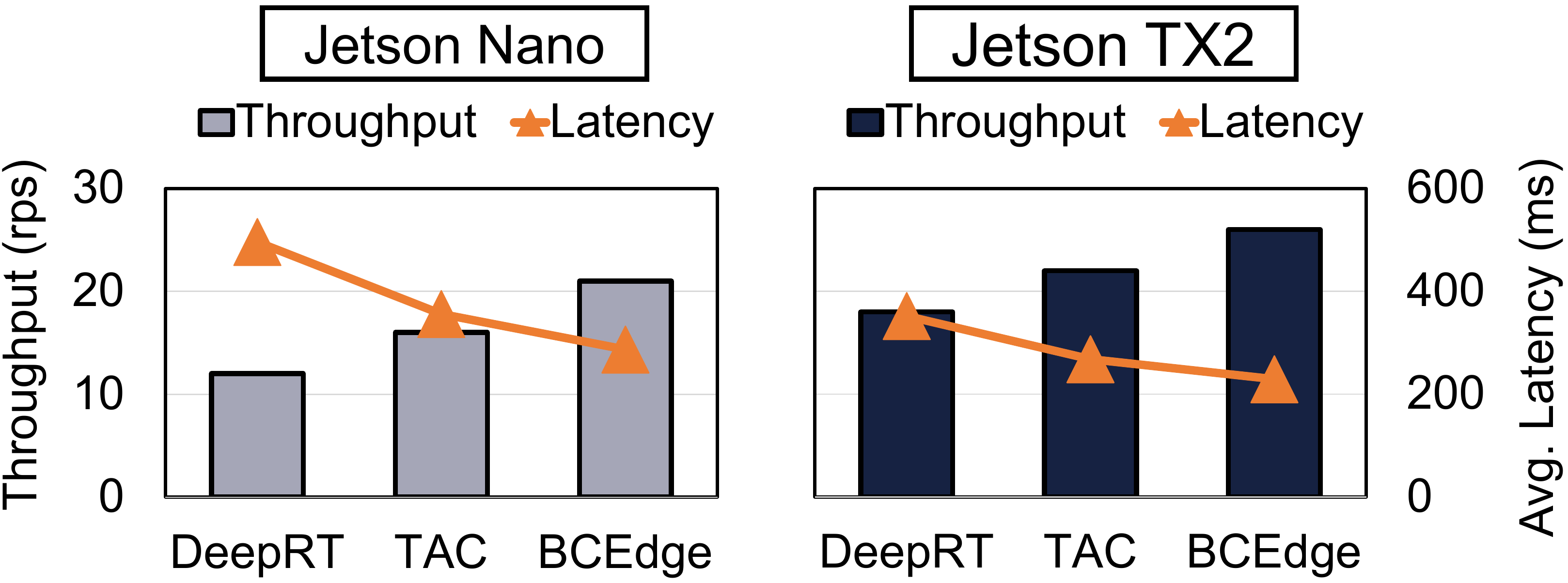}
\end{minipage}}
\subfigure[ResNet-18 (res)]{
\begin{minipage}[b]{\linewidth}
\includegraphics[width=1\linewidth]{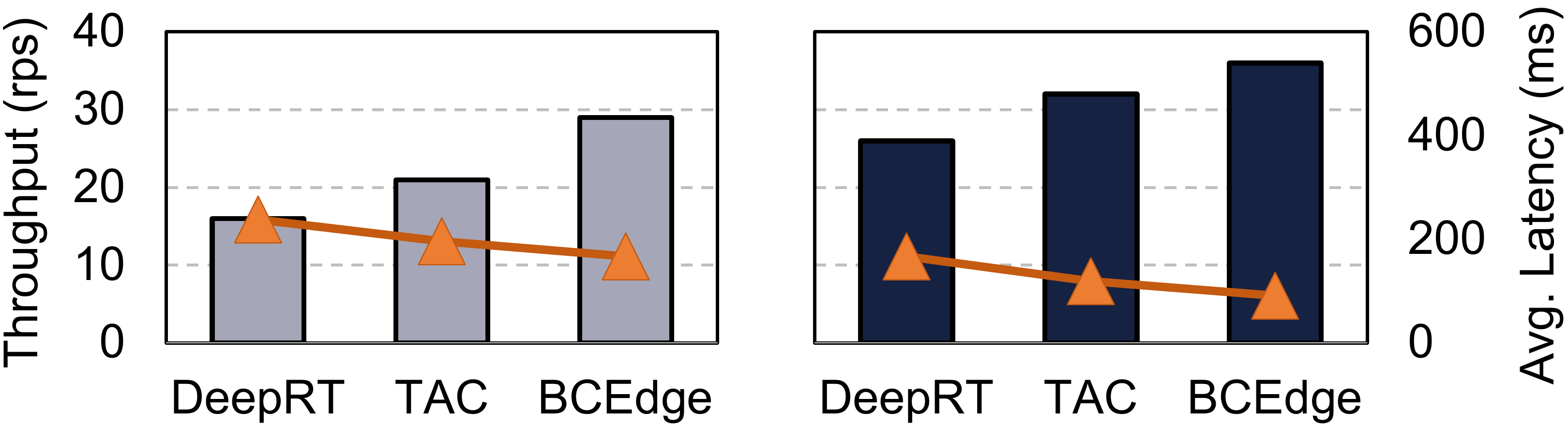}
\end{minipage}}
\subfigure[TinyBERT (bert)]{
\begin{minipage}[b]{\linewidth}
\includegraphics[width=1\linewidth]{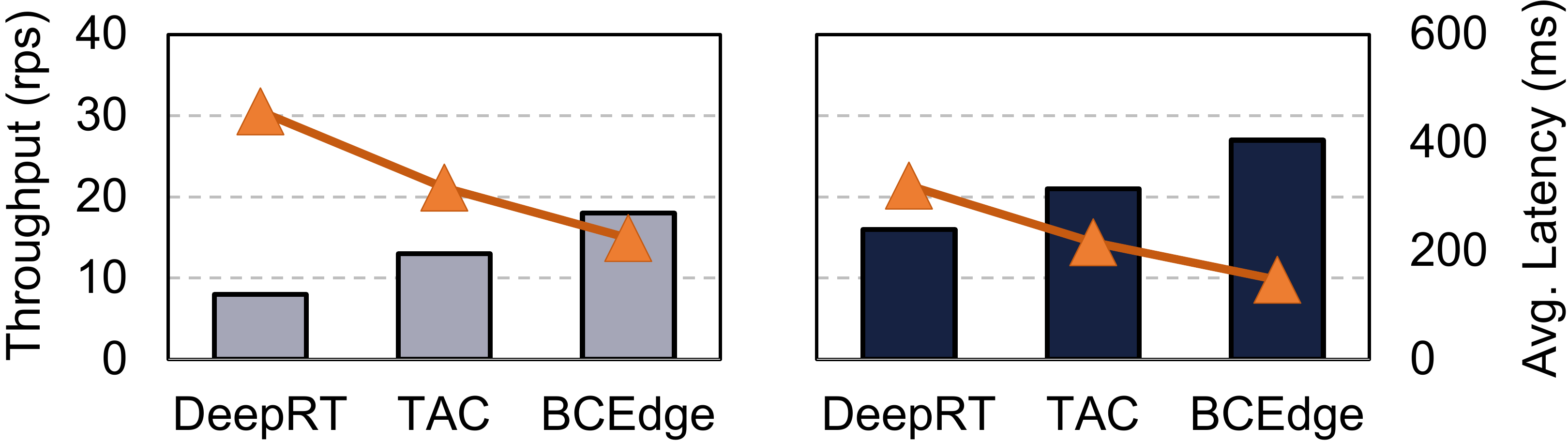}
\end{minipage}}
\caption{The throughput and average latency of heterogeneous edge platforms. The more computing resources of the edge platform, the higher the throughput and the lower the average latency.}
\label{fig:result}
\end{figure}

Fig.~\ref{fig:result} shows the throughput and latency of three edge platforms corresponding to the utility in Fig.~\ref{fig:device_utility}. 
The left y-axis in Fig.~\ref{fig:result} represents peak throughput, and the right y-axis represents corresponding average latency.
As we observed in Fig.~\ref{fig:result}, BCEdge also has more significant performance improvement on the DNN models with fewer computing resources and the edge platforms with more abundant computing resources.
Even for the weakest Jetson Nano, BCEdge can fully utilize computing resources for different DNN models to optimize the trade-off between throughput and latency.
In summary, BCEdge exhibits flexible scalability that can be adapted to heterogeneous resource-constrained edge platforms.

\subsection{Evaluation of Interference Model}
\begin{figure}[htbp]
\large
\centerline{\includegraphics[width=0.7\linewidth]{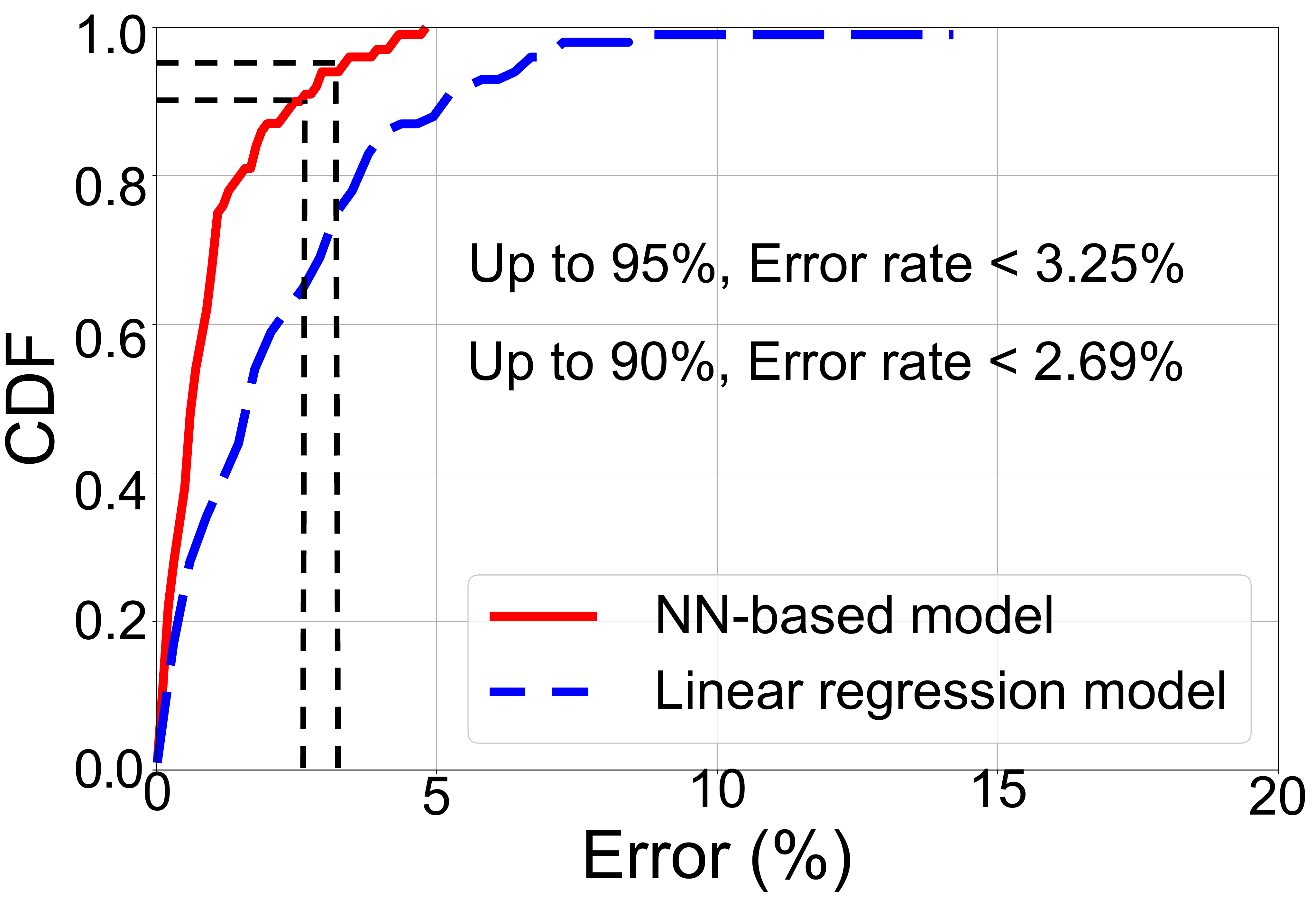}}
\caption{Cumulative distribution of relative error rate. Our proposed NN-based model can predict up-to 95\% of cases with less than 3.25\% error rate.}
\label{fig:CDF_Pre}
\vspace{-0.25cm} 
\end{figure}

In this experiment, we investigate the proposed interference prediction model in BCEdge under different requests per second (rps) on SLO violation rate. 
The interference prediction model records total 2000 inference interferences with one second period for each DNN model. Among the 2000 pieces of collected data, we randomly select 1600 pieces of execution data as training data and 400 pieces of data for validation.
Fig.~\ref{fig:CDF_Pre} presents the cumulative distribution of the prediction error with our NN-based interference model, compared with the linear regression model~\cite{chen2016baymax,choi2022serving}. 
The proposed model can predict up to 90\% of cases within 2.69\% error rate and up to 95\% if 3.25\% of error is allowed, which reduces the error rate by half compared to the linear regression model.
Since the model interference we observed in Fig.~\ref{observation} is not a simple linear relationship, the linear regression model has a higher prediction error. 
In contrast, out proposed NN-based interference model considers the resource utilization of edge platforms and the actual latency of DNN models, which can accurately predict the interference latency.

\begin{figure}[htbp]
\large
\centerline{\includegraphics[width=0.7\linewidth]{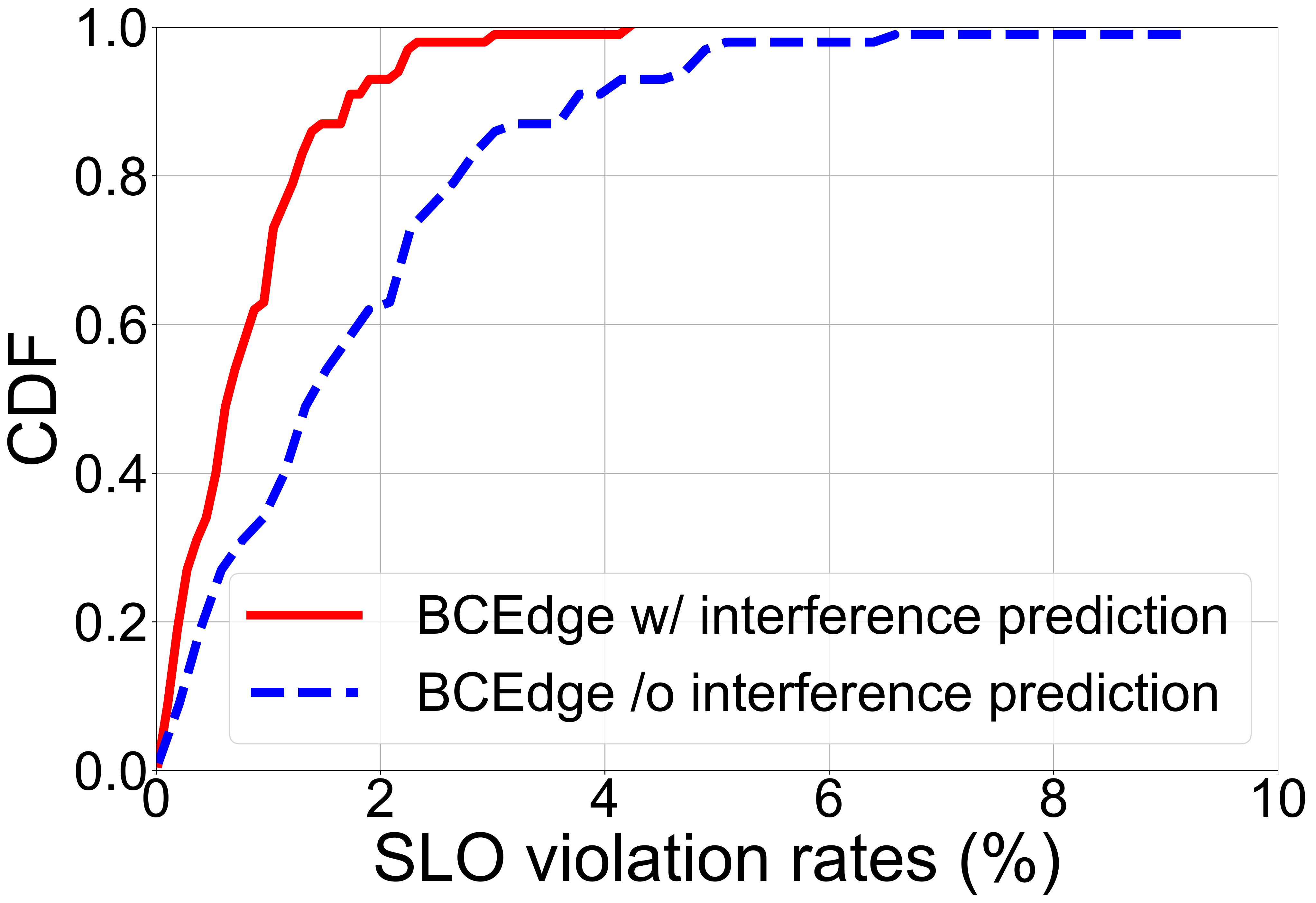}}
\caption{Cumulative distribution of SLO violation rate with 30rps. 
Our proposed interference prediction model can achieve the SLO violation rate within 4\%, compared to up to 9.2\% SLO violation rate without the interference prediction model.}
\label{fig:predict}
\vspace{-0.25cm} 
\end{figure}

Fig.~\ref{fig:predict} shows the cumulative distribution of SLO violation rate at 30 rps for BCEdge with/without the interference prediction model.
We analyzed the SLO violation rate for the scheduling duration of 3000 seconds in Fig.~\ref{fig:throughput} and Fig.~\ref{fig:latency}.
The proposed model can reduce the SLO violation rate of BCEdge from 9.2\% to 4.1\%, compared to BCEdge without the interference prediction model.
It illustrates that the interference prediction model can improve the robustness of BCEdge and effectively reduce SLO violation rate.

\begin{figure}[htbp]
\centerline{\includegraphics[width=\linewidth]{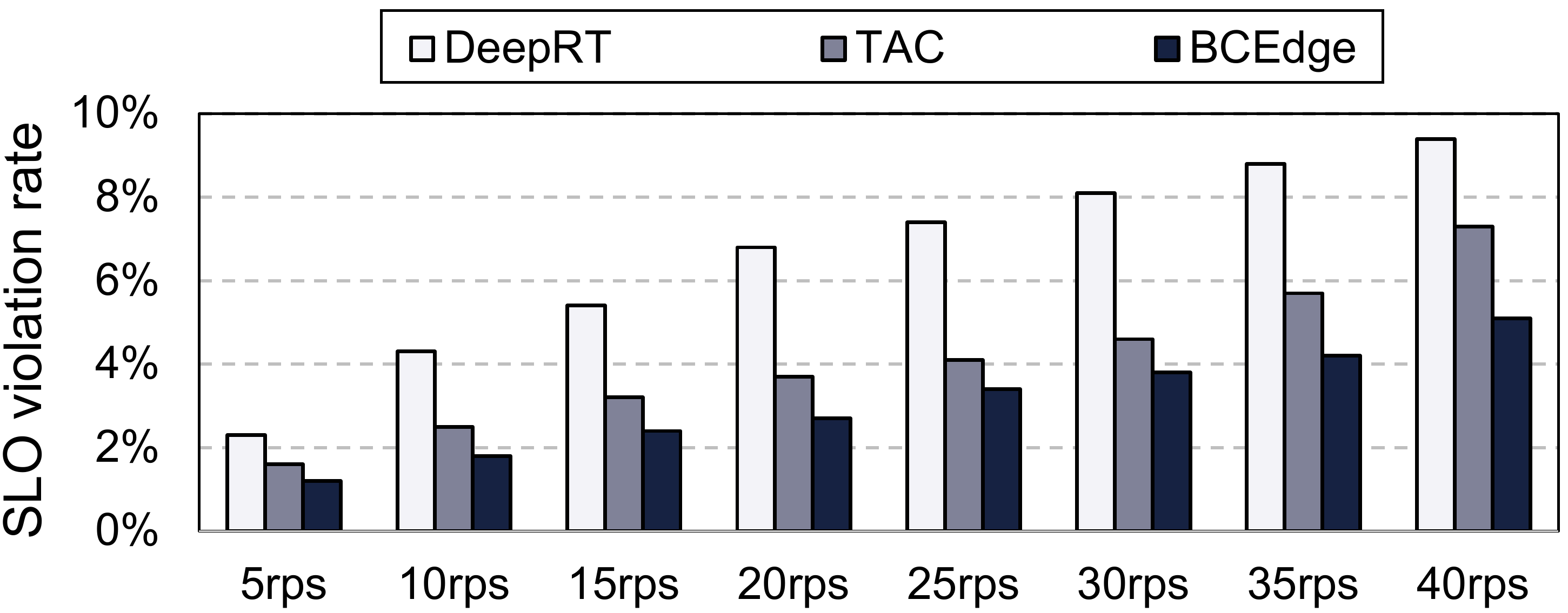}}
\caption{Comparison of service level objective (SLO) violation rate with real-world six DNN model benchmarks under different rps. Benefit from the interference prediction model, BCEdge has the lowest SLO violation rate.}
\label{fig:slo}
\vspace{-0.25cm} 
\end{figure}

As shown in Fig.~\ref{fig:slo}, we measure the SLO violation rate by gradually increasing the requests per second (rps). 
It can be seen that BCEdge has the lowest SLO violation rate for all rps, which is 53\% and 25\% lower than DeepRT and TAC on average, respectively. 
The SLO violation rate of BCEdge does not exceed 5\% even at 40rps.
Since the soft real-time scheduler in DeepRT~\cite{yang2021deeprt} is only suitable for the DNN models without strict SLO constraints, it has the highest SLO violation rate with strict SLO constraints. 
In addition, TAC does not consider the impact of model interference, therefore the SLO violation rate is higher than BCEdge.

\subsection{Scheduling Overhead}
\begin{figure}[htbp]
\centerline{\includegraphics[width=\linewidth]{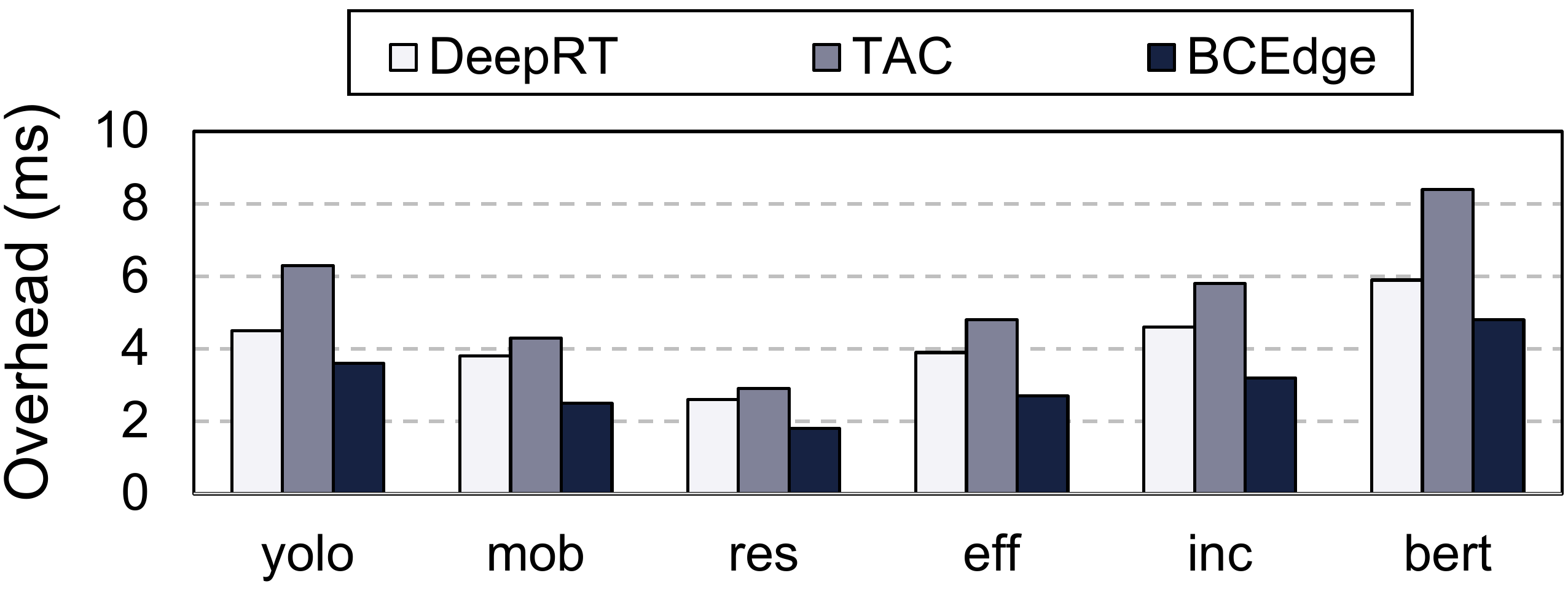}}
\caption{Comparison of scheduling overhead with six DNN models.}
\label{fig:cost}
\vspace{-0.2cm}
\end{figure}

To measure the runtime overhead imposed by the scheduler, we compare BCEdge with DeepRT and TAC in terms of the average scheduling latency. 
Fig.~\ref{fig:cost} depicts these scheduling overheads. 
As observed, BCEdge has a low scheduling overhead due to the scheduler in BCEdge introduces maximum entropy that can learn more approximate optimal actions to speed up learning in order to reduce overhead.
Specifically, the average scheduling overhead of BCEdge is 26\% and 43\% lower than that of DeepRT and TAC, respectively. 
It demonstrates that BCEdge can efficiently schedule batching and concurrent requests with extremely low overhead.
Note that we did not evaluate the overhead of the performance profiler and interference prediction model as they are negligible.

\section{Conclusion}\label{Conclusion}
In this work, we present BCEdge, an adaptive, SLO-aware, and multi-tenant DNN-serving scheduling framework.
BCEdge enables batching and concurrent inference for edge intelligent applications on edge platforms
to achieve both high-throughput and low-latency.
The key to BCEdge is a maximum entropy deep reinforcement learning-based scheduler, which automatically co-optimizes batch size and number of concurrent models.
Compared to the state-of-the-art solutions, BCEdge achieves up to 37.6\% average utility improvement, while satisfying SLOs.

{\small
\bibliographystyle{IEEEtran}
\bibliography{ref}

\begin{thebibliography}{10}
\providecommand{\url}[1]{#1}
\csname url@samestyle\endcsname
\providecommand{\newblock}{\relax}
\providecommand{\bibinfo}[2]{#2}
\providecommand{\BIBentrySTDinterwordspacing}{\spaceskip=0pt\relax}
\providecommand{\BIBentryALTinterwordstretchfactor}{4}
\providecommand{\BIBentryALTinterwordspacing}{\spaceskip=\fontdimen2\font plus
\BIBentryALTinterwordstretchfactor\fontdimen3\font minus
  \fontdimen4\font\relax}
\providecommand{\BIBforeignlanguage}[2]{{%
\expandafter\ifx\csname l@#1\endcsname\relax
\typeout{** WARNING: IEEEtran.bst: No hyphenation pattern has been}%
\typeout{** loaded for the language `#1'. Using the pattern for}%
\typeout{** the default language instead.}%
\else
\language=\csname l@#1\endcsname
\fi
#2}}
\providecommand{\BIBdecl}{\relax}
\BIBdecl

\bibitem{crankshaw2017clipper}
D.~Crankshaw, X.~Wang, G.~Zhou, M.~J. Franklin, J.~E. Gonzalez, and I.~Stoica,
  ``Clipper: A low-latency online prediction serving system.'' in \emph{NSDI},
  vol.~17, 2017, pp. 613--627.

\bibitem{olston2017tensorflow}
C.~Olston, N.~Fiedel, K.~Gorovoy, J.~Harmsen, L.~Lao, F.~Li, V.~Rajashekhar,
  S.~Ramesh, and J.~Soyke, ``Tensorflow-serving: Flexible, high-performance ml
  serving,'' \emph{arXiv preprint arXiv:1712.06139}, 2017.

\bibitem{Triton}
``Nvidia triton inference server,''
  \url{https://developer.nvidia.com/nvidia-triton-inference-server}.

\bibitem{deng2020edge}
S.~Deng, H.~Zhao, W.~Fang, J.~Yin, S.~Dustdar, and A.~Y. Zomaya, ``Edge
  intelligence: The confluence of edge computing and artificial intelligence,''
  \emph{IEEE Internet of Things Journal}, vol.~7, no.~8, pp. 7457--7469, 2020.

\bibitem{feng2020deep}
D.~Feng, C.~Haase-Sch{\"u}tz, L.~Rosenbaum, H.~Hertlein, C.~Glaeser, F.~Timm,
  W.~Wiesbeck, and K.~Dietmayer, ``Deep multi-modal object detection and
  semantic segmentation for autonomous driving: Datasets, methods, and
  challenges,'' \emph{IEEE Transactions on Intelligent Transportation Systems},
  vol.~22, no.~3, pp. 1341--1360, 2020.

\bibitem{xie2022contrastive}
X.~Xie, F.~Sun, Z.~Liu, S.~Wu, J.~Gao, J.~Zhang, B.~Ding, and B.~Cui,
  ``Contrastive learning for sequential recommendation,'' in \emph{2022 IEEE
  38th international conference on data engineering (ICDE)}.\hskip 1em plus
  0.5em minus 0.4em\relax IEEE, 2022, pp. 1259--1273.

\bibitem{liu2020collabar}
Z.~Liu, G.~Lan, J.~Stojkovic, Y.~Zhang, C.~Joe-Wong, and M.~Gorlatova,
  ``Collabar: Edge-assisted collaborative image recognition for mobile
  augmented reality,'' in \emph{2020 19th ACM/IEEE International Conference on
  Information Processing in Sensor Networks (IPSN)}.\hskip 1em plus 0.5em minus
  0.4em\relax IEEE, 2020, pp. 301--312.

\bibitem{han2015deep}
S.~Han, H.~Mao, and W.~J. Dally, ``Deep compression: Compressing deep neural
  networks with pruning, trained quantization and huffman coding,'' \emph{arXiv
  preprint arXiv:1510.00149}, 2015.

\bibitem{YOLOv5}
``Yolov5,'' \url{https://github.com/ultralytics/yolov5}.

\bibitem{TensorRT}
``Nvidia tensorrt,'' \url{https://developer.nvidia.com/tensorrt}.

\bibitem{tuli2020dynamic}
S.~Tuli, S.~Ilager, K.~Ramamohanarao, and R.~Buyya, ``Dynamic scheduling for
  stochastic edge-cloud computing environments using a3c learning and residual
  recurrent neural networks,'' \emph{IEEE transactions on mobile computing},
  2020.

\bibitem{yang2021deeprt}
Z.~Yang, K.~Nahrstedt, H.~Guo, and Q.~Zhou, ``Deeprt: A soft real time
  scheduler for computer vision applications on the edge,'' in \emph{2021
  IEEE/ACM Symposium on Edge Computing (SEC)}.\hskip 1em plus 0.5em minus
  0.4em\relax IEEE, 2021, pp. 271--284.

\bibitem{choi2020prema}
Y.~Choi and M.~Rhu, ``Prema: A predictive multi-task scheduling algorithm for
  preemptible neural processing units,'' in \emph{2020 IEEE International
  Symposium on High Performance Computer Architecture (HPCA)}.\hskip 1em plus
  0.5em minus 0.4em\relax IEEE, 2020, pp. 220--233.

\bibitem{cui2022dvabatch}
W.~Cui, H.~Zhao, Q.~Chen, H.~Wei, Z.~Li, D.~Zeng, C.~Li, and M.~Guo,
  ``$\{$DVABatch$\}$: Diversity-aware $\{$Multi-Entry$\}$$\{$Multi-Exit$\}$
  batching for efficient processing of $\{$DNN$\}$ services on $\{$GPUs$\}$,''
  in \emph{2022 USENIX Annual Technical Conference (USENIX ATC 22)}, 2022, pp.
  183--198.

\bibitem{zhang2019mark}
C.~Zhang, M.~Yu, W.~Wang, and F.~Yan, ``Mark: Exploiting cloud services for
  cost-effective, slo-aware machine learning inference serving.'' in
  \emph{USENIX Annual Technical Conference}, 2019, pp. 1049--1062.

\bibitem{choi2022serving}
S.~Choi, S.~Lee, Y.~Kim, J.~Park, Y.~Kwon, and J.~Huh, ``Serving heterogeneous
  machine learning models on $\{$Multi-GPU$\}$ servers with
  $\{$Spatio-Temporal$\}$ sharing,'' in \emph{2022 USENIX Annual Technical
  Conference (USENIX ATC 22)}, 2022, pp. 199--216.

\bibitem{seo2021slo}
W.~Seo, S.~Cha, Y.~Kim, J.~Huh, and J.~Park, ``Slo-aware inference scheduler
  for heterogeneous processors in edge platforms,'' \emph{ACM Transactions on
  Architecture and Code Optimization (TACO)}, vol.~18, no.~4, pp. 1--26, 2021.

\bibitem{romero2021infaas}
F.~Romero, Q.~Li, N.~J. Yadwadkar, and C.~Kozyrakis, ``Infaas: Automated
  model-less inference serving.'' in \emph{USENIX Annual Technical Conference},
  2021, pp. 397--411.

\bibitem{258862}
A.~Gujarati, R.~Karimi, S.~Alzayat, W.~Hao, A.~Kaufmann, Y.~Vigfusson, and
  J.~Mace, ``Serving {DNNs} like clockwork: Performance predictability from the
  bottom up,'' in \emph{14th USENIX Symposium on Operating Systems Design and
  Implementation (OSDI 20)}, 2020, pp. 443--462.

\bibitem{ali2020batch}
A.~Ali, R.~Pinciroli, F.~Yan, and E.~Smirni, ``Batch: Machine learning
  inference serving on serverless platforms with adaptive batching,'' in
  \emph{SC20: International Conference for High Performance Computing,
  Networking, Storage and Analysis}.\hskip 1em plus 0.5em minus 0.4em\relax
  IEEE, 2020, pp. 1--15.

\bibitem{liu2021leia}
X.~Liu, B.~Wu, X.~Yuan, and X.~Yi, ``Leia: A lightweight cryptographic neural
  network inference system at the edge,'' \emph{IEEE Transactions on
  Information Forensics and Security}, vol.~17, pp. 237--252, 2021.

\bibitem{hou2021model}
J.~Hou, H.~Liu, Y.~Liu, Y.~Wang, P.-J. Wan, and X.-Y. Li, ``Model protection:
  Real-time privacy-preserving inference service for model privacy at the
  edge,'' \emph{IEEE Transactions on Dependable and Secure Computing}, 2021.

\bibitem{kim2020autoscale}
Y.~G. Kim and C.-J. Wu, ``Autoscale: Energy efficiency optimization for
  stochastic edge inference using reinforcement learning,'' in \emph{2020 53rd
  Annual IEEE/ACM International Symposium on Microarchitecture (MICRO)}.\hskip
  1em plus 0.5em minus 0.4em\relax IEEE, 2020, pp. 1082--1096.

\bibitem{zhang2021deep}
W.~Zhang, D.~Yang, H.~Peng, W.~Wu, W.~Quan, H.~Zhang, and X.~Shen, ``Deep
  reinforcement learning based resource management for dnn inference in
  industrial iot,'' \emph{IEEE Transactions on Vehicular Technology}, vol.~70,
  no.~8, pp. 7605--7618, 2021.

\bibitem{han2022microsecond}
M.~Han, H.~Zhang, R.~Chen, and H.~Chen, ``Microsecond-scale preemption for
  concurrent $\{$GPU-accelerated$\}$$\{$DNN$\}$ inferences,'' in \emph{16th
  USENIX Symposium on Operating Systems Design and Implementation (OSDI 22)},
  2022, pp. 539--558.

\bibitem{cui2021enable}
W.~Cui, H.~Zhao, Q.~Chen, N.~Zheng, J.~Leng, J.~Zhao, Z.~Song, T.~Ma, Y.~Yang,
  C.~Li \emph{et~al.}, ``Enable simultaneous dnn services based on
  deterministic operator overlap and precise latency prediction,'' in
  \emph{Proceedings of the International Conference for High Performance
  Computing, Networking, Storage and Analysis}, 2021, pp. 1--15.

\bibitem{liu2022veltair}
Z.~Liu, J.~Leng, Z.~Zhang, Q.~Chen, C.~Li, and M.~Guo, ``Veltair: towards
  high-performance multi-tenant deep learning services via adaptive compilation
  and scheduling,'' in \emph{Proceedings of the 27th ACM International
  Conference on Architectural Support for Programming Languages and Operating
  Systems}, 2022, pp. 388--401.

\bibitem{mondal2021scheduling}
S.~S. Mondal, N.~Sheoran, and S.~Mitra, ``Scheduling of time-varying workloads
  using reinforcement learning,'' in \emph{Proceedings of the AAAI Conference
  on Artificial Intelligence}, vol.~35, no.~10, 2021, pp. 9000--9008.

\bibitem{han2021tailored}
Y.~Han, S.~Shen, X.~Wang, S.~Wang, and V.~C. Leung, ``Tailored learning-based
  scheduling for kubernetes-oriented edge-cloud system,'' in \emph{IEEE INFOCOM
  2021-IEEE Conference on Computer Communications}.\hskip 1em plus 0.5em minus
  0.4em\relax IEEE, 2021, pp. 1--10.

\bibitem{wang2020multi}
X.~Wang, Z.~Ning, and S.~Guo, ``Multi-agent imitation learning for pervasive
  edge computing: A decentralized computation offloading algorithm,''
  \emph{IEEE Transactions on Parallel and Distributed Systems}, vol.~32, no.~2,
  pp. 411--425, 2020.

\bibitem{zou2020a3c}
J.~Zou, T.~Hao, C.~Yu, and H.~Jin, ``A3c-do: A regional resource scheduling
  framework based on deep reinforcement learning in edge scenario,'' \emph{IEEE
  Transactions on Computers}, vol.~70, no.~2, pp. 228--239, 2020.

\bibitem{shi2020mean}
D.~Shi, H.~Gao, L.~Wang, M.~Pan, Z.~Han, and H.~V. Poor, ``Mean field game
  guided deep reinforcement learning for task placement in cooperative
  multiaccess edge computing,'' \emph{IEEE Internet Things Journal}, vol.~7,
  no.~10, pp. 9330--9340, 2020.

\bibitem{tuli2021mcds}
S.~Tuli, G.~Casale, and N.~R. Jennings, ``Mcds: Ai augmented workflow
  scheduling in mobile edge cloud computing systems,'' \emph{IEEE Transactions
  on Parallel and Distributed Systems}, vol.~33, no.~11, pp. 2794--2807, 2021.

\bibitem{meng2019online}
J.~Meng, H.~Tan, X.-Y. Li, Z.~Han, and B.~Li, ``Online deadline-aware task
  dispatching and scheduling in edge computing,'' \emph{IEEE Transactions on
  Parallel and Distributed Systems}, vol.~31, no.~6, pp. 1270--1286, 2019.

\bibitem{qiao2021pollux}
A.~Qiao, S.~K. Choe, S.~J. Subramanya, W.~Neiswanger, Q.~Ho, H.~Zhang, G.~R.
  Ganger, and E.~P. Xing, ``Pollux: Co-adaptive cluster scheduling for
  goodput-optimized deep learning,'' in \emph{15th $\{$USENIX$\}$ Symposium on
  Operating Systems Design and Implementation ($\{$OSDI$\}$ 21)}, 2021.

\bibitem{christodoulou2019soft}
P.~Christodoulou, ``Soft actor-critic for discrete action settings,''
  \emph{arXiv preprint arXiv:1910.07207}, 2019.

\bibitem{haarnoja2018soft}
T.~Haarnoja, A.~Zhou, K.~Hartikainen, G.~Tucker, S.~Ha, J.~Tan, V.~Kumar,
  H.~Zhu, A.~Gupta, P.~Abbeel \emph{et~al.}, ``Soft actor-critic algorithms and
  applications,'' \emph{arXiv preprint arXiv:1812.05905}, 2018.

\bibitem{howard2017mobilenets}
A.~G. Howard, M.~Zhu, B.~Chen, D.~Kalenichenko, W.~Wang, T.~Weyand,
  M.~Andreetto, and H.~Adam, ``Mobilenets: Efficient convolutional neural
  networks for mobile vision applications,'' \emph{arXiv preprint
  arXiv:1704.04861}, 2017.

\bibitem{he2016deep}
K.~He, X.~Zhang, S.~Ren, and J.~Sun, ``Deep residual learning for image
  recognition,'' in \emph{Proceedings of the IEEE conference on computer vision
  and pattern recognition}, 2016, pp. 770--778.

\bibitem{tan2019efficientnet}
M.~Tan and Q.~Le, ``Efficientnet: Rethinking model scaling for convolutional
  neural networks,'' in \emph{International conference on machine
  learning}.\hskip 1em plus 0.5em minus 0.4em\relax PMLR, 2019, pp. 6105--6114.

\bibitem{szegedy2016rethinking}
C.~Szegedy, V.~Vanhoucke, S.~Ioffe, J.~Shlens, and Z.~Wojna, ``Rethinking the
  inception architecture for computer vision,'' in \emph{Proceedings of the
  IEEE conference on computer vision and pattern recognition}, 2016, pp.
  2818--2826.

\bibitem{jiao-etal-2020-tinybert}
X.~Jiao, Y.~Yin, L.~Shang, X.~Jiang, X.~Chen, L.~Li, F.~Wang, and Q.~Liu,
  ``{T}iny{BERT}: Distilling {BERT} for natural language understanding,'' in
  \emph{Findings of the Association for Computational Linguistics: EMNLP 2020},
  Nov. 2020, pp. 4163--4174.

\bibitem{whitley1994genetic}
D.~Whitley, ``A genetic algorithm tutorial,'' \emph{Statistics and computing},
  vol.~4, no.~2, pp. 65--85, 1994.

\bibitem{schulman2017proximal}
J.~Schulman, F.~Wolski, P.~Dhariwal, A.~Radford, and O.~Klimov, ``Proximal
  policy optimization algorithms,'' \emph{arXiv preprint arXiv:1707.06347},
  2017.

\bibitem{van2016deep}
H.~Van~Hasselt, A.~Guez, and D.~Silver, ``Deep reinforcement learning with
  double q-learning,'' in \emph{Proceedings of the AAAI conference on
  artificial intelligence}, vol.~30, no.~1, 2016.

\bibitem{chen2016baymax}
Q.~Chen, H.~Yang, J.~Mars, and L.~Tang, ``Baymax: Qos awareness and increased
  utilization for non-preemptive accelerators in warehouse scale computers,''
  \emph{ACM SIGPLAN Notices}, vol.~51, no.~4, pp. 681--696, 2016.

\end{thebibliography}
}
\vspace{-15mm}
\begin{IEEEbiography}[{\includegraphics[width=1in,height=1.25in,clip,keepaspectratio]{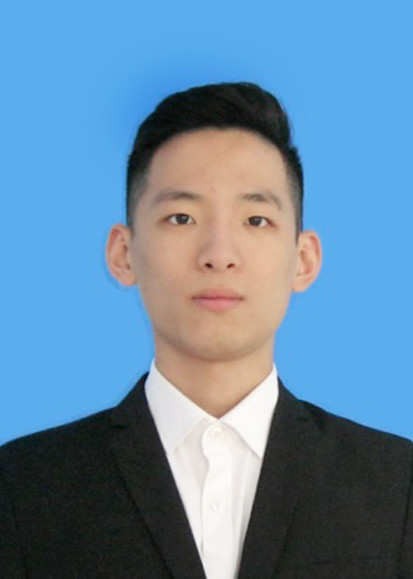}}]
{Ziyang Zhang}
Ziyang Zhang received the MS degrees in the School of Electronic Information and Optical Engineering, Nankai University, Tianjin, China, in 2020.
He is currently working toward the PhD degree in the School of Computer Science and Technology, Harbin Institute of Technology (HIT), Harbin, China.
His research interests include edge computing, machine learning system, and deep learning.
\end{IEEEbiography}
\vspace{-15mm}
\begin{IEEEbiography}[{\includegraphics[width=1in,height=1.25in,clip,keepaspectratio]{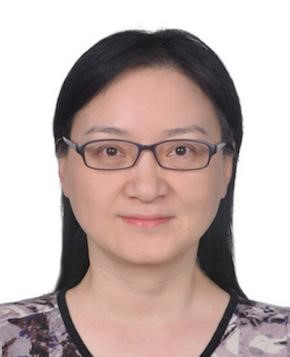}}]
{Huan Li}
Dr. Huan Li obtained her PhD degree in Computer Science from the University of Massachusetts at Amherst, USA in 2006. 
Her current research interests include AIoT, Edge intelligence, distributed real-time systems, and data science. 
She has served as program committee member for numerous international conferences including IEEE RTAS, ICDCS, RTCSA, etc. 
She is now a senior member of IEEE. 
\end{IEEEbiography}
\vspace{-15mm}
\begin{IEEEbiography}[{\includegraphics[width=1in,height=1.25in,clip,keepaspectratio]{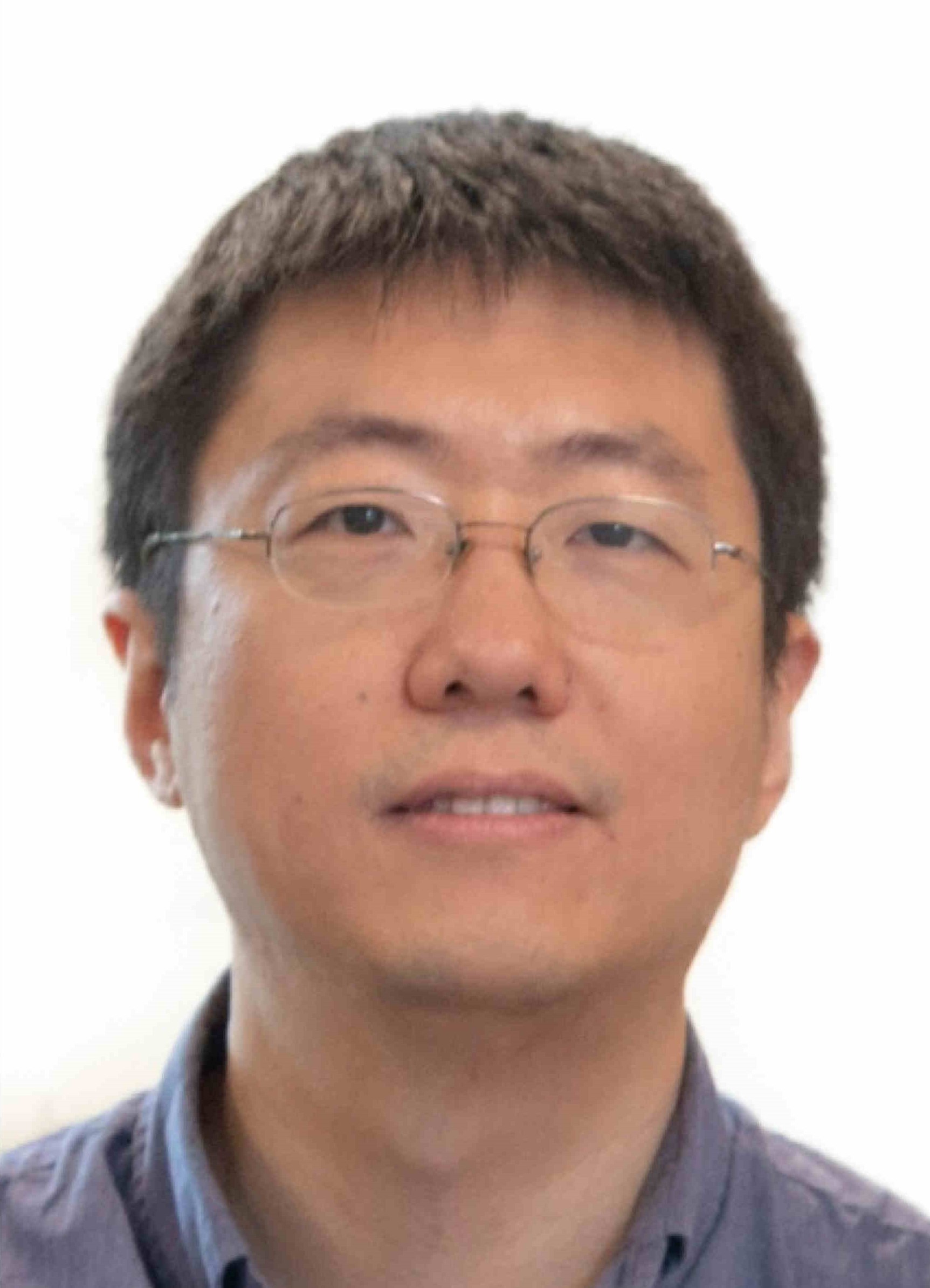}}]
{Yang Zhao}
Yang Zhao received the BS degree (2003) in electrical engineering from Shandong University, the MS degree (2006) in electrical engineering from the Beijing University of Aeronautics and Astronautics, and the PhD degree (2012) in electrical and computer engineering from the University of Utah. He was a lead research engineer at GE Global Research between 2013 and 2021. Since 2021, he has been at Harbin Institute of Technology, Shenzhen, where he is a research professor in the International Research Institute for Artificial Intelligence. His research interests include wireless sensing, edge computing and cyber physical systems. He is a senior member of the IEEE. 
\end{IEEEbiography}
\vspace{-125mm}
\begin{IEEEbiography}[{\includegraphics[width=1in,height=1.25in,clip,keepaspectratio]{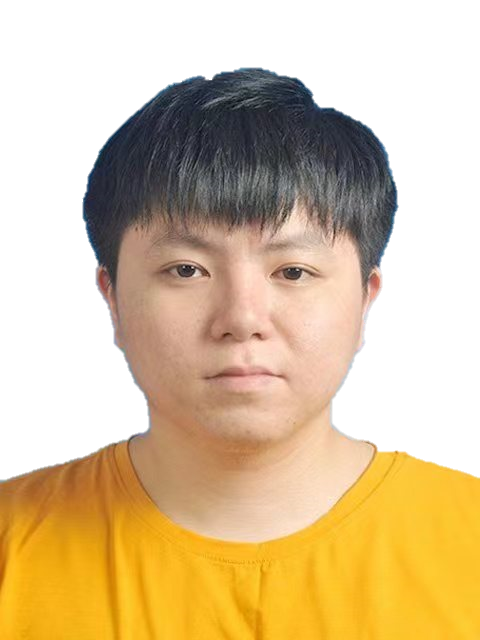}}]
{Changyao Lin}
Changyao Lin received the BS and MS degrees in the School of Computer Science and Technology, Harbin Institute of Technology (HIT), Harbin, China, in 2020 and 2022, respectively. 
He is currently working toward the PhD degree at HIT. 
His research interests include edge computing, distributed system, and deep learning.
\end{IEEEbiography} 
\vspace{-125mm}
\begin{IEEEbiography}[{\includegraphics[width=1in,height=1.25in,clip,keepaspectratio]{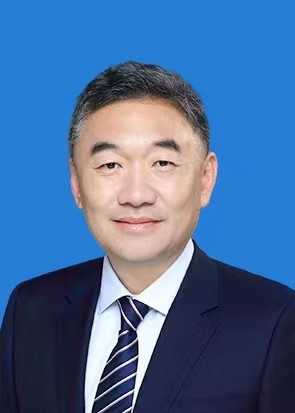}}]
{Jie Liu}
Jie Liu is a Chair Professor at Harbin Institute of Technology Shenzhen (HIT Shenzhen), China and the Dean of its AI Research Institute. 
Before joining HIT, he spent 18 years at Xerox PARC and Microsoft. 
He was a Principal Research Manager at Microsoft Research, Redmond and a partner of the company. 
His research interests are Cyber-Physical Systems, AI for IoT, and energy efficient computing. 
He received IEEE TCCPS Distinguished Leadership Award and 6 Best Paper Awards from top conferences. 
He is an IEEE Fellow and an ACM Distinguished Scientist, and founding Chair of ACM SIGBED China.
\end{IEEEbiography}

\end{document}